\documentclass[a4paper,fleqn]{cas-dc}

\usepackage[authoryear]{natbib}

\usepackage{booktabs}
\usepackage{colortbl}
\usepackage{amssymb}
\usepackage{amsmath}
\usepackage[table]{xcolor}

\usepackage{tikz}

\newcommand{\bentarrow}[1][]{%
  \begin{tikzpicture}[#1]%
    \draw (0,0.7ex) -- (0,0) -- (0.75em,0);
    \draw (0.55em,0.2em) -- (0.75em,0) -- (0.55em,-0.2em);
  \end{tikzpicture}%
}

\usepackage{soul}
\usepackage{xcolor}
\definecolor{lightgray}{gray}{0.25} 
\setstcolor{lightgray}

\usepackage{pdflscape}
\usepackage{afterpage}
\usepackage{capt-of}

\definecolor{high}{HTML}{65f6ab}
\definecolor{medium}{HTML}{F3DE8A}
\definecolor{low}{HTML}{EB9486}
\definecolor{no}{HTML}{ADACAF}
\definecolor{temporal}{HTML}{94A3DB}
\definecolor{demographic}{HTML}{8CBCB9}

\usepackage{listings}
\usepackage{xcolor}
\usepackage{upquote}      
\lstdefinestyle{python}{
    language=Python,
    basicstyle=\ttfamily\small,
    keywordstyle=\color{blue},
    commentstyle=\color{gray},
    stringstyle=\color{teal},
    numbers=left,
    numberstyle=\tiny,
    stepnumber=1,
    numbersep=8pt,
    frame=single,
    breaklines=true,
    showstringspaces=false
}

\usepackage{tcolorbox}
\usepackage{fvextra}

\usepackage[dvipsnames]{xcolor}

\def\tsc#1{\csdef{#1}{\textsc{\lowercase{#1}}\xspace}}
\tsc{WGM}
\tsc{QE}

\begin{document}
\let\WriteBookmarks\relax
\def\floatpagepagefraction{1}
\def\textpagefraction{.001}

\shorttitle{}    

\shortauthors{}  

\title [mode = title]{CXRMate-2: Structured Multimodal Temporal Embeddings and Tractable Reinforcement Learning for Clinically Acceptable Chest X-ray Radiology Report Generation}  



%


\author[1]{Aaron Nicolson}[orcid=0000-0002-7163-1809]

\cormark[1]


\ead{aaron.nicolson@csiro.au}

\author[1]{Elizabeth J. Cooper}
\author[1]{Hwan-Jin Yoon}
\author[1]{Claire McCafferty}



\affiliation[1]{organization={Australian e-Health Research Centre, CSIRO Health and Biosecurity},
            city={Brisbane \& Melbourne},
            country={Australia}}





\author[2]{Ramya Krishnan}
\author[2]{Michelle Craigie}
\author[2]{Nivene Saad}

\affiliation[2]{organization={Princess Alexandra Hospital, Metro South Health},
            city={Brisbane},
            country={Australia}}

\author[1]{Jason Dowling}
\author[3,4]{Ian A. Scott}

\affiliation[3]{organization={Queensland Digital Health Centre, University of Queensland},
            city={Brisbane},
            country={Australia}}

\affiliation[4]{organization={Digital Health and Informatics Directorate, Metro South Hospital and Health Service},
            city={Brisbane},
            country={Australia}}

\author[1,5]{Bevan Koopman}

\affiliation[5]{organization={School of Electrical Engineering and Computer Science, University of Queensland},
            city={Brisbane},
            country={Australia}}

\cortext[1]{Corresponding author}



\begin{abstract}
Chest X-ray (CXR) radiology report generation (RRG) models have shown rapid progress on automated metrics, yet their clinical utility remains uncertain due to limited qualitative evaluation by radiologists. We present CXRMate-2, a state-of-the-art CXR RRG model that enables tractable reinforcement learning (RL) through structured multimodal temporal embeddings and high-resolution visual feature compression, for efficient, unified conditioning of an LLM decoder on visual, textual, and temporal context from a study and its prior. This enables group relative policy optimisation (GRPO), where a proposed reward function is used to improve semantic alignment with radiologist reports. Across the MIMIC-CXR, CheXpert Plus, and ReXgradient datasets, CXRMate-2 achieves statistically significant improvements over strong benchmarks, including gains of 11.2\% and 24.4\% in GREEN and RadGraph-XL, respectively, on MIMIC-CXR relative to MedGemma 1.5 (4B).

To directly compare CXRMate-2 against radiologist reporting, we conduct a blinded, randomised qualitative retrospective evaluation. Three consultant radiologists compare generated and radiologist reports across 120 studies from the MIMIC-CXR test set. Generated reports were deemed acceptable---defined as preferred or rated equally to radiologist reports---in 45\% of ratings, with no statistically significant difference in preference rates for seven of the eight analysed findings. Preferences for radiologist reports were driven primarily by higher recall, while generated reports were consistently preferred for readability.

Together, these results define a clear pathway to clinically acceptable CXR RRG. Improving recall and the detection of subtle findings represents the primary remaining barrier to non-inferiority with radiologist reporting, positioning CXR RRG for prospective evaluation in assistive, radiologist-led workflows.

\end{abstract}





\maketitle

\section{Introduction} 

Chest X-rays (CXRs) are a rapid, low-cost, low-radiation modality used for initial thoracic assessment, accounting for an estimated 40\% of the 3.5–4 billion imaging studies performed annually \citep{dasegowda_suboptimal_2023}. Rising imaging volumes and workforce shortages have led to reporting backlogs, delays, and radiologist burnout, undermining timely patient care~\citep{bailey_understanding_2022, singh_occupational_2017}.

Vision--language artificial intelligence (AI) models specialised for CXR radiology report generation (RRG) have emerged as a promising tool to support radiologists, with the capability to automatically generate reports from images and patient data \citep{bannur_maira-2_2024, nicolson_impact_2025}. Recent CXR RRG systems combine elements such as a self-supervised vision encoder pre-trained on large-scale CXR datasets; a decoder conditioned on the textual and visual information from a study and its prior; a large language model (LLM) as the decoder; training on multiple large-scale publicly-available CXR RRG datasets; and multi-stage fine-tuning, with reinforcement learning (RL) using rewards for semantic alignment with radiologist reports as the final stage. In practice, however, integration remains difficult due to the substantial computational complexity of unifying these components—further exacerbated by reinforcement learning algorithms such as grouped relative policy optimisation (GRPO)—with most models therefore incorporating only a subset of these elements. As a result, the potential benefits of these components are not fully realised when used in isolation, thereby limiting overall model performance.
 
Despite rapid progress in model capability, evaluation practices have not kept pace. Most studies continue to rely primarily on automated metrics that quantify similarity between generated and radiologist reports. These metrics exhibit only moderate correlation with radiologist judgement and therefore have limited capacity to assess findings, impressions, and recommendations of generated reports \citep{ostmeier_green_2024, zhao_ratescore_2024}. Qualitative evaluation by radiologists is therefore necessary to properly assess these models and support their safe translation into practice. However, few studies conduct rigorous qualitative evaluations of radiologist preferences, and even fewer analyse the underlying reasons for preference, statistical power, inter-rater agreement, or the effects of different factors.

In this work, we address limitations in both model design and qualitative evaluation for CXR RRG. First, we propose CXRMate-2, a state-of-the-art (SOTA) model that unifies key components for CXR RRG within a single framework. In parallel, we conduct a qualitative retrospective evaluation with consultant radiologists, using blinded pairwise comparisons to assess report preferences, capture the underlying reasons for these preferences, and perform comprehensive statistical analysis. The remainder of the paper is organised around these two components: \textbf{CXRMate-2 for chest X-ray radiology report generation} and its \textbf{qualitative retrospective evaluation}.

\paragraph{\textbf{CXRMate-2 for chest X-ray (CXR) radiology report generation (RRG):}} 
CXRMate-2 integrates RAD-DINO as the visual encoder \citep{perez-garcia_exploring_2025}; a query-based Transformer adapter (Q-Adapter) for efficient high-resolution visual feature compression; an LLM decoder conditioned on visual, textual, and temporal information from a study and its prior via structured multimodal temporal embeddings; GRPO with a proposed composite reward function; and training on three large-scale publicly-available CXR datasets: MIMIC-CXR-JPG, CheXpert Plus, and ReXgradient. The full methodology for CXRMate-2 is described in Subsection \ref{sec:meth_cxrmate2}, with results in Subsection \ref{sec:results_cxrmate2}. Its key findings and contributions are:

 
\begin{itemize}
    \item Achieves SOTA performance across multiple public datasets and automated metrics, with statistically significant improvements over strong baselines, including improvements of 11.2\% and 24.4\% in GREEN and RadGraph-XL, respectively, on MIMIC-CXR relative to MedGemma 1.5 (4B) \citep{sellergren_medgemma_2026}.
    \item Tractable GRPO enabled by reducing self-attention computational complexity using structured multimodal temporal embeddings and a query-based Transformer adapter (Subsection \ref{sec:architecture}).
    \item Proposed composite reward for RL that improves semantic alignment with radiologist reports (Subsection \ref{sec:rl_results}).
    \item Improves generalisation by training on three large-scale CXR RRG public datasets (Subsection \ref{sec:datasets}).
    \item Releases model checkpoints (\url{https://huggingface.co/aehrc/cxrmate-2}) and training code (\url{https://github.com/aehrc/cxrmate-2}) for reproducibility.
\end{itemize}

\paragraph{\textbf{Qualitative retrospective evaluation:}}
We conducted a qualitative retrospective evaluation to explore radiologist preferences between generated and radiologist reports using a blinded, randomised pairwise comparison design with a SOTA CXR RRG model. Three consultant radiologists independently evaluated 120 studies from the MIMIC-CXR test set. For each study, radiologist raters were shown two reports---one radiologist-written and one model-generated---and asked to select their preferred report and indicate the reason(s) for the preference (precision, recall, and/or readability). Secondary analyses evaluated statistical power, inter-rater agreement, and the effects of different factors. The methodology for the qualitative retrospective evaluation is described in Subsection \ref{sec:retro_meth}, with results presented and discussed in Subsection \ref{sec:results_retro}. The key findings and contributions include:


\begin{itemize}
    \item Generated reports were deemed acceptable---defined as preferred or equally preferred  to radiologist reports---in 45\% of ratings (Subsection \ref{sec:results_pref}).
    \item Eight findings were targeted for the evaluation, including atelectasis, cardiomegaly, pulmonary edema, normal/no findings, pneumonia, pulmonary congestion, simple pleural effusion, and simple pneumothorax.
    \item No statistically significant difference was observed between acceptable generated reports and radiologist reports for seven of the eight targeted findings; the exception was pulmonary congestion.
    \item Recall was the primary driver of preference for radiologist reports. They more consistently captured a larger share of findings, impressions, and recommendations, whereas generated reports were more often preferred for readability in abnormal studies (Subsection \ref{sec:results_reasons}).
    \item There was low inter-rater agreement, with a mean pairwise agreement of 50\% and Fleiss’ $\kappa = 0.16$, indicating only slight agreement beyond chance (Subsection \ref{sec:inter}).
    \item It was found that several factors significantly influenced generated report acceptability, including the reason, the rater, the finding, and the interaction between the rater and the reason (Subsection \ref{sec:effects}).
\end{itemize}

Taken together, this work provides a unified view of current model capability and key limitations relative to radiologists---particularly in recall and in handling subtle, diffuse findings such as pulmonary congestion---thereby delineating concrete directions for achieving non-inferiority to radiologist reports and, ultimately, clinically acceptable CXR RRG. 

If non-inferiority to radiologist reports is achieved, CXR RRG models are appropriately positioned as assistive tools within radiologist-led workflows. Possibilities include use as a first-reader to generate draft reports, as a second-reader to support quality assurance, for automated triage based on study severity, and as predictive text to accelerate report writing. These use cases align with the strengths of current models---particularly readability and efficiency---while mitigating risks associated with incomplete or imperfect findings, impressions, and recommendations.

\section{Related work}

\subsection{Qualitative evaluation of chest X-ray (CXR) radiology report generation (RRG)}

Qualitative evaluation of CXR RRG remains limited by methodological shortcomings in previous work, including small sample sizes, limited statistical analysis, and a lack of insight into the factors driving radiologist preferences. As a result, conclusions based solely on automated metrics---or on existing qualitative studies---may not reflect the requirements for integration into radiologist workflows.

\cite{miura_improving_2021} proposed a reward for RL that promotes improved factual completeness and consistency between generated and radiologist reports. \cite{delbrouck_improving_2022} similarly introduced the RadGraph reward, which improves the factual correctness between generated and radiologist reports \citep{jain_radgraph_2021}. In both studies, models trained with these entity-based rewards were preferred over comparison models in radiologist pairwise evaluations on 100 MIMIC-CXR test studies. However, the evaluations did not quantify the performance gap between generated and radiologist reports.

Several studies have qualitatively evaluated generated reports versus radiologist reports, but often with limited methodological rigour. In \citet{huang_generative_2023}, generated reports achieved similar Likert scores to radiologist reports and outperformed teleradiology in emergency department (ED) CXRs ($n\mathord{=}500$); however, the cohort was dominated by normal studies (67.2\%), and no analysis of preference drivers or statistical power was provided. \citet{chen_chexagent_2024} found that radiologist raters preferred radiologist reports over CheXagent-generated reports across completeness, correctness, and conciseness, though the evaluation was limited to 20 studies. For MedVersa, \citet{zhou_medversa_2026} reported low preference rates for generated reports (7\%) and high indifference (64\%), but did not include statistical testing or analyses of inter-rater agreement and influencing factors. Similarly, \citet{tanno_collaboration_2025} observed variable preference for generated reports across cohorts (28.8\% US; 57.7\% India), but did not report reasons for preferences, statistical power, or inter-rater agreement.

A series of studies from the same group evaluated KARA-CXR, a proprietary model trained on 8.8 million CXR–report pairs. In \citet{hong_value_2025}, five radiologists interpreted CXRs with and without KARA-CXR as a first-reader; the use of KARA-CXR was associated with reduced reading times (34.2 to 19.8 seconds), improved sensitivity for selected findings, and increased inter-rater agreement. In \citet{hwang_clinical_2025}, seven radiologists judged 64.1\% of generated reports ($n\mathord{=}1\,539$) to be acceptable without modification, supporting its use as a first-reader. However, a survey study by \citet{jeong_artificial_2025} found that radiologists remained cautious, expressing neutrality regarding accuracy and endorsing use primarily for screening, while opposing stand-alone deployment without further validation.

In contrast, this work presents a qualitative retrospective evaluation using a blinded, randomised pairwise comparison between generated and radiologist reports across 120 studies from the MIMIC-CXR test set, assessed by consultant radiologists. Beyond report preferences, we analyse the reasons underlying those preferences and perform comprehensive statistical evaluation, including inter-rater agreement, factors associated with preferences, statistical power, and estimation of sample sizes required for future studies.

\subsection{Recent developments in chest X-ray (CXR) radiology report generation (RRG)}

Incorporating additional patient context has improved performance on automated metrics in CXR RRG. Early work leveraged multiple CXRs per study, since CXR examinations often include complementary frontal and lateral views of the patient \citep{miura_improving_2021, gaber_lateral_2005}. Further gains were found by incorporating a patient’s prior study to identify changes over time \citep{nicolson_longitudinal_2024, wu_deltanet_2022, kelly_chest_2012, bannur_learning_2023, liu_recap_2023}. Additional performance was achieved by including the \textit{indication} and \textit{history} sections of the report as input, as well as structured electronic health record (EHR) data from the emergency department for ED cases \citep{nguyen_pragmatic_2023, nicolson_impact_2025}. Following these approaches, our model CXRMate-2 takes as input the indication, history, comparison, and technique sections from the radiologist report of a study, all CXRs of the study and its prior, and the findings and impression section from the radiology report of the prior study. Overall, these advances suggest that richer patient context is an important driver of performance in CXR RRG, which we leverage through the comprehensive set of inputs used by CXRMate-2.

MAIRA-1 provided another leap in performance, by leveraging a self-supervised visual encoder pre-trained on large-scale CXR datasets, specifically RAD-DINO, and an instruction-tuned LLM as the decoder \citep{hyland_maira-1_2024}. The visual encoder was frozen and its features mapped for the input of the LLM via an adapter, which was four-layer feedforward neural network (FNN). MAIRA-2 expanded upon this by increasing the training data through the inclusion of multiple public and private sources, forming a large-scale training dataset \citep{bannur_maira-2_2024}. Like MAIRA-2, CXRMate-2 utilises RAD-DINO as its visual encoder, an LLM as its decoder, and an improved query-based Transformer adapter for efficient high-resolution visual feature compression. It is also trained on a large-scale dataset comprising three publicly available sources: MIMIC-CXR-JPG, CheXpert Plus, and ReXgradient. Collectively, these extensions build upon MAIRA-1 and MAIRA-2 by introducing efficient high-resolution visual feature compression for improved multimodal conditioning, while relying exclusively on public datasets to ensure reproducibility.

\begin{figure*}[]
  \centering
  \includegraphics[width=\textwidth]{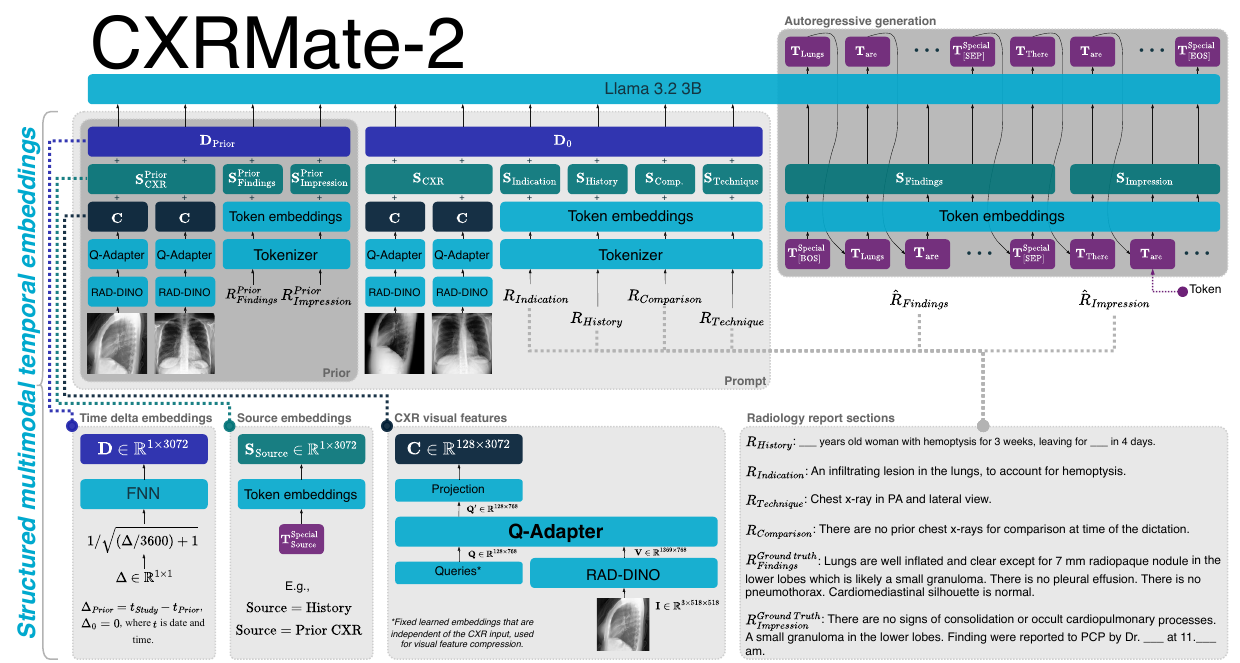}
  \caption{CXRMate-2 for CXR RRG; its rich clinical context spans both the study and its prior. Inputs include all available CXRs from both timepoints; report sections from the study (indication, history, comparison, and technique); findings and impression from the prior report; and the time delta between them. This is efficiently provided to the LLM decoder via structured multimodal temporal embeddings that integrate visual features with token,  source, and time-delta embeddings, enabling unified modelling of visual, textual, and temporal information. This representation reduces self-attention computational complexity, enabling tractable optimisation with GRPO. Here, an improved composite reward comprising RaTEScore, CXR-BERT, BERTScore, and ARN increases semantic alignment with radiologist reports. $\mathbf{T}\in \{0,1\}^{128\,256}$ and $\|\mathbf{T}\|_1=1$.}
  \label{fig:model}

  \includegraphics[width=\textwidth]{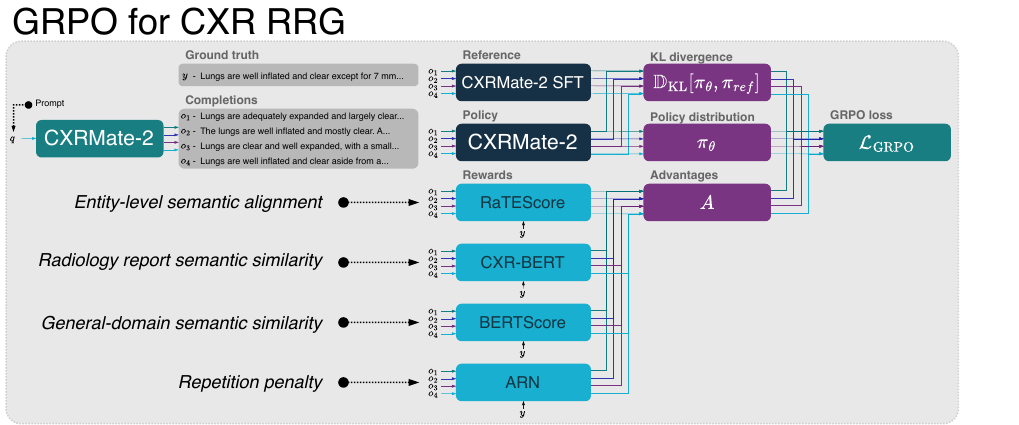}
  \caption{GRPO for CXR RRG. The final stage of fine-tuning for CXRMate-2 constitutes GRPO with our proposed composite reward function comprising RaTEScore, CXR-BERT, BERTScore, and ARN, which increases semantic alignment with radiologist reports. CXRMate-2 SFT refers to the model obtained after supervised fine-tuning, before to the GRPO stage.}
  \label{fig:grpo}
\end{figure*}

An alternative to instruction prompting used by models such as MAIRA-2 is the structured multimodal temporal embeddings introduced by our previous model, CXRMate \citep{nicolson_longitudinal_2024}. Here, learned embeddings are added to the input so that the decoder can distinguish heterogeneous data sources without relying on natural-language instructions. This reduces input size and self-attention computational complexity, and may mitigate attention dilution \citep{qin_devil_2022}. Our previous model CXRMate-ED expanded upon this by adding time delta embeddings, which encode the time difference between the study and preceding events, such as the prior study \citep{nicolson_impact_2025}. CXRMate-2 replaces instruction prompting with structured multimodal temporal embeddings to reduce input length and self-attention computational complexity, enabling more efficient training with resource-intensive RL algorithms such as GRPO.


RL has been increasingly used to improve semantic alignment between generated and radiologist reports. \cite{liu_clinically_2019} introduced self-critical sequence training (SCST) with rewards capturing readability and semantic alignment with radiologist reports, with subsequent work improving both optimisation methods and reward design \citep{nicolson_e-health_2024, nicolson_impact_2025}. More recently, group relative policy optimisation (GRPO) has been applied to CXR RRG, offering improved training stability and reward optimisation \citep{lin_toward_2026}. These developments reflect a shift toward directly optimising report-level semantic fidelity rather than relying solely on likelihood-based training. CXRMate-2 builds on this by applying GRPO with a composite reward that encourages semantic alignment with radiologist reports.


Recent generalist medical imaging foundation models support multiple vision–language tasks. MedVersa is a unified multimodal LLM trained across diverse modalities and clinical text, while MedGemma leverages large-scale medical pre-training and instruction-tuning to achieve competitive CXR RRG performance despite not being task-specific \citep{zhou_medversa_2026, sellergren_medgemma_2025}. These models reflect a shift from specialised systems toward multi-task approaches driven by data scale and instruction alignment.


Most existing work advances CXR RRG along individual axes. In contrast, CXRMate-2 integrates these complementary components within a unified framework, combining comprehensive visual, textual, and temporal context, a self-supervised visual encoder with efficient query-based feature compression, a large language model (LLM) decoder, structured multimodal temporal embeddings for scalable conditioning, large-scale publicly-available multi-institutional training data, and GRPO with an improved composite reward for improved semantic alignment with radiologist reports. This holistic design not only consolidates existing advances but also addresses their practical limitations---particularly computational scalability and optimisation stability---enabling more effective end-to-end training and improved alignment with radiologist reports.

\section{CXRMate-2 for chest X-ray (CXR) radiology report generation (RRG)} \label{sec:cxrmate2_section}

This section presents CXRMate-2, our SOTA CXR RRG model, and its automated quantitative evaluation. We assess its performance across multiple public benchmarks. In the methodology (Section \ref{sec:meth_cxrmate2}), we describe the model architecture, including the Q-Adapter for efficient visual feature compression and the structured multimodal temporal embedding framework for visual, textual, and temporal conditioning, as well as the training data and multi-stage training procedure, including RL with GRPO and the proposed composite reward. GRPO is not tractable for this task under modest compute constraints without the Q-Adapter and structured multimodal temporal embeddings, making these components essential for fine-tuning CXRMate-2 with GRPO. We then present results using a broad suite of automated metrics, alongside analyses that quantify the contributions of the key architectural and training design choices. A qualitative evaluation  of CXRMate-2 by radiologist raters is presented in the next section (Section \ref{sec:qual_section}).

\subsection{Methodology} \label{sec:meth_cxrmate2}

\subsubsection{Model architecture} \label{sec:architecture}

CXRMate-2 is illustrated in Figure \ref{fig:model}; it unifies and extends the design principles of CXRMate-ED and MAIRA-2, incorporating architectural, prompting, and RL improvements over both models.

CXRMate-2 inherits the structured prompt embeddings of CXRMate-ED. This is an alternative strategy to instruction-based prompting where different data sources are indicated to the decoder via adding source embeddings. A source embedding is a learned embedding added to each input to encode which predefined input type it belongs to, enabling the decoder to differentiate heterogeneous inputs without relying on natural-language instructions. This substantially reduces the number of inputs passed to the decoder by eliminating the natural-language instructional scaffolding that would otherwise surround each input type. 

CXRMate-2 also inherits the time-delta embeddings from CXRMate-ED, which are used to inject temporal information into the prompt inputs. Finally, it generates the findings and impression sections in the same manner; by generating a special separator token, the two sections can be extracted from the generated tokens. It also inherits the non-causal (bidirectional) attention masking over the prompt tokens, while maintaining causal masking for generated tokens.

CXRMate-2 also improves upon the RL employed by CXRMate-ED. First, we replace SCST \citep{rennie_self-critical_2017} with GRPO \citep{shao_deepseekmath_2024}---a more stable algorithm that samples multiple completions per prompt, computes rewards relative to the group of completions, and constrains policy updates via a divergence penalty to a reference model. The composite reward of CXRMate-ED is further augmented with RaTEScore, a report-level evaluation metric shown to correlate with radiologist quality assessments \citep{zhao_ratescore_2024}.

From MAIRA-2, CXRMate-2 adopts its RAD-DINO visual encoder. Specifically, we use the RAD-DINO checkpoint released with MAIRA-2 as the visual backbone, which is kept frozen throughout training.\footnote{\url{https://huggingface.co/microsoft/rad-dino-maira-2}} We improve upon MAIRA-2 by replacing the Vicuna 7B v1.5 decoder with LLaMA 3.2 3B, which was found to improve performance despite having fewer parameters. 

Furthermore, we replace its four-layer FNN adapter---that maps the CXR visual features into the token embedding latent space---with the Q-Adapter. The Q-Adapter is described in detail in Subsection \ref{sec:q_adapter}; it uses multiple Transformer encoder layers and a set of learnable queries to reduce the number of CXR visual features for the decoder. By reducing the number of visual tokens passed to the decoder, it decreases the self-attention sequence length, substantially lowering self-attention computational complexity, and consequently reduces the model’s VRAM requirements. This reduction was important for enabling tractable GRPO, which is memory-intensive due to maintaining a reference model and sampling multiple completions per prompt. This is further discussed in Subsection \ref{sec:results_cxrmate2}.

CXRMate-2 is trained on three publicly-available datasets: MIMIC-CXR-JPG, CheXpert Plus, and ReXgradient, as described in Subsection \ref{sec:datasets}. These datasets provide diverse institutional distributions, reporting styles, and patient populations, improving generalisation across clinical settings.

Results for each of these improvements are presented in  Subsections \ref{sec:development}--\ref{sec:datasets}.

\subsubsection{Q-Adapter} \label{sec:q_adapter}

The Q-Adapter transforms the CXR visual features into a fixed set of token-aligned embeddings to be given as input to the decoder. It replaces the four-layer FNN adapter used by MAIRA-2. It is inspired by query resampling architectures such as the Perceiver Resampler \citep{alayrac_flamingo_2022} and operates by introducing a set of learnable latent query embeddings that attend to visual features through a stack of Transformer encoder layers, as shown in Figure \ref{fig:model}. The queries are fixed learned parameters that do not depend on the input CXR.

Visual features $\mathbf{V} \in \mathbb{R}^{1369 \times 768}$ are concatenated with a set of learned queries $\mathbf{Q} \in \mathbb{R}^{128 \times 768}$ to form a joint input. This is processed by multiple RoFormer encoder layers \citep{su_roformer_2024}---two for CXRMate-2---that perform self-attention over both the queries and visual features, allowing the queries to aggregate global contextual information from the visual representation. After the final layer, only the outputs corresponding to the query tokens are retained and projected through a linear layer to match the hidden dimensionality of the decoder, producing a compact representation $\mathbf{C} \in \mathbb{R}^{128 \times 3072}$.

\subsubsection{Structured multimodal temporal embeddings}

CXRMate-2 inherits the structured multimodal temporal embeddings of CXRMate-ED, which itself extends the original CXRMate embedding design. While CXRMate-ED paired CXR studies with structured EHR tables from MIMIC-IV-ED, we do not incorporate EHR tables here and focus exclusively on imaging and report-related inputs. The exclusion of structured ED EHR tables is a limitation of this work, as they contain complementary clinical context (e.g., vital signs, medications) that has been shown to improve CXR RRG performance on ED studies. Time delta and source embeddings are added to both token embeddings and projected visual features before being passed to the decoder.

\paragraph{\textbf{Source embeddings:}} A source embedding is a learned embedding added to each input to encode which predefined input type it belongs to. In practice, we reuse unused special token embeddings from the LLM vocabulary as source embeddings. In total, CXRMate-2 employs ten source embeddings; six for each of the report sections, two for the prior findings and impression sections, one for the CXRs of the study, and one for the CXRs of the prior. 

\paragraph{\textbf{Time delta embeddings:}} Temporal information is added to the prompt through time delta embeddings. In our setting, only the date and time for a study and its prior are considered. The reference is the study date and time; therefore, the time delta for inputs from the study is zero. For the prior, the time delta is the difference between the date and time of the study and the prior.

To construct the time-delta embeddings, the time difference $\Delta$ (in seconds) is first transformed as $\Delta' = (\Delta/3\,600+1)^{-1/2}$. This yields a value of 1.0 when $\Delta = 0$ and monotonically decreases toward zero as the time delta increases, with diminishing sensitivity. The transformation therefore assigns greater weight to events that are temporally closer to the study date and time, effectively prioritising more recent events under the assumption that they are more clinically relevant, although this may not hold in all cases (e.g., interval assessment of lung nodules). $\Delta'$ is then projected to the decoder hidden size using a two-layer FNN $\textbf{D} = f(\Delta'\textbf{W}_1)\textbf{W}_2$, where $\textbf{W}_1 \in \mathbb{R}^{1 \times 3\,072}$, $\textbf{W}_2 \in \mathbb{R}^{3\,072 \times 2\,048}$, and $f(\cdot)$ is the GELU activation function \citep{hendrycks_gaussian_2023}. The resultant embedding is added to the corresponding token or visual feature representations.

\paragraph{\textbf{Position identifier ordering:}} Position identifiers follow the ordering used in CXRMate-ED. Inputs are first sorted by time delta, such that smaller time deltas are positioned closer to generated tokens, and then by CXR view, such that frontal views are positioned closer to the generated tokens than lateral views. This ordering exploits the properties of rotary positional embeddings (RoPE) \citep{su_roformer_2024} used by the decoder, which bias attention toward nearby tokens, thus encouraging stronger interactions between them.

\subsubsection{Training and evaluation data}

We used a combination of large, publicly-available collections of de-identified CXR studies for model training and evaluation, including MIMIC-CXR-JPG \citep{johnson_mimic-cxr_2019,johnson_mimic-cxr-jpg_2024}, CheXpert Plus \citep{chambon_chexpert_2024}, and ReXgradient \citep{zhang_rexgradient-160k_2025}, each associated with a structured radiology report. MIMIC-CXR-JPG is the JPG version of the MIMIC-CXR dataset, CheXpert Plus comprises studies acquired at Stanford Health Care, and ReXgradient consists of studies collected across multiple US health systems. A summary of the number of patients, studies, and CXRs in each dataset is provided in Table \ref{tab:dataset_summary}.

\begin{table}[t]
\centering
\setlength{\tabcolsep}{6pt}
\small
\rowcolors{2}{gray!15}{white}
\caption{\label{tab:dataset_summary}Summary of datasets used for training and evaluation.}
\begin{tabular}{lccc}
\toprule
Dataset & Patients & Studies & CXRs \\
\midrule
MIMIC-CXR-JPG & 65\,379 & 227\,827 & 377\,110 \\
CheXpert Plus & 64\,725 & 187\,711 & 223\,462 \\
ReXGradient-160K & 109\,487 & 160\,000 & 273\,004 \\
\bottomrule
\end{tabular}
\end{table}

Training was executed with the the official training splits of each dataset; validation was executed with the official MIMIC-CXR-JPG validation split; and testing was executed with the official test split of MIMIC-CXR-JPG, the validation split of CheXpert Plus, and the official public test split of ReXgradient. The number of studies in the training, validation, and test splits are reported in each table. Each image from each dataset was normalised using Listing \ref{lst:mimic_cxr_norm}, to standardise intensity ranges and enhance local contrast across images from heterogeneous sources, thereby reducing variability arising from differences in acquisition protocols and post-processing. After normalisation, each CXR was processed with the transforms of the RAD-DINO image processor.

\subsubsection{Training}

CXRMate-2 was trained using supervised fine-tuning (SFT) followed by GRPO. Optimisation was performed with \textit{AdamW}~\citep{loshchilov_decoupled_2019} using mini-batch gradient descent, with the following hyperparameters: $\beta_1=0.9$, $\beta_2=0.999$, $\epsilon=1e-8$, and $\lambda=0.01$. Training was conducted on $8\times$ AMD Instinct MI300X 192GB GPUs with FP32 precision. Training data were restricted to studies with radiologist reports containing both findings and impression section; this requirement was not applied to prior studies. After filtering, the training set comprised $313\,503$ studies and the validation set contained $991$ studies. 

During training, up to five CXRs were provided as input per example. If the total number of CXRs from a given study and its prior exceeded five, five CXRs were sampled uniformly at random for that example, and the remainder were not used for that forward pass. The validation RaTEScore was the monitored metric for checkpoint selection.

A maximum of 320 generated tokens was set for validation, testing, and for generating completions for GRPO. Greedy search decoding was used for validation and testing.

\paragraph{\textbf{SFT:}}
SFT was performed with a local mini-batch size of 2 (resulting in a global mini-batch size of 16) and a maximum of five epochs. Only RAD-DINO was frozen during this stage.  We trained the model using standard token-level cross-entropy loss for next-token prediction. We used a cosine learning rate schedule with linear warm-up (500 steps) and hard restarts, comprising five cycles over five epochs (one per epoch), with a peak learning rate of 5e-5.

\paragraph{\textbf{GRPO:}}
RL with GRPO was performed after SFT, as shown in Figure \ref{fig:grpo}. The reward components comprised RaTEScore, CXR-BERT, BERTScore, and absence of repeated $n$-grams (ARN) \citep[Subsection D.1]{nicolson_impact_2025}. The reward weights were 0.3 for RaTEScore, CXR-BERT, and BERTScore, and 0.1 for ARN. GRPO was performed with a local mini-batch size of 1 (resulting in a global mini-batch size of 8) and a maximum of two epochs.

We used a constant learning rate of 1e-6 with a linear warm-up of 500 training steps. Only the decoder was not frozen during this stage. For each prompt $q$, we sample a group of $G$ completions $\{o_i\}_{i=1}^{G}$ from the current policy $\pi_\theta$. When sampling completions, top-$k$ and top-$p$ sampling were not applied and a temperature of 1.0 was set. 

Let $T_i$ denote the number of completion tokens for sample $i$, and let $m_{i,t} \in \{0,1\}$ be a mask indicating valid completion tokens. Padding tokens, along with completions that did not include exactly one beginning-of-sentence, separator, and end-of-sentence special token were masked.

We compute a composite reward for each completion, and convert rewards into group-normalised advantages. Concretely, for each reward component $k$ we compute rewards $r_{i,k}$ and form a normalised advantage over the $k^{th}$ component:
\[
a_{i,k}=\frac{r_{i,k} - \mu_k}{\sigma_k},~{\rm where}
\]
\[
\quad\mu_k=\frac{1}{G}\sum_{i=1}^{G} r_{i,k}~{\rm and}~\sigma_k = \sqrt{\frac{1}{G}\sum_{i=1}^{G} (r_{i,k}-\mu_k)^2}.
\]
The advantage for the $i^{th}$ completion is then:
\[
A_i = \sum_k w_k\, a_{i,k}.
\]
where $w_k$ is the weight for the $k$-th reward component. The GRPO loss is then:
\[
\begin{aligned}
\mathcal{L}_{\mathrm{GRPO}}(\theta)
&=
-\frac{1}{
\sum_{i=1}^{G}\sum_{t=1}^{T_i} m_{i,t}
}
\sum_{i=1}^{G}\sum_{t=1}^{T_i}
m_{i,t}
\Big[
 \\
&\min\big(
\rho_{i,t}(\theta) A_i,\tilde{\rho}_{i,t}(\theta) A_i
\big)\\
&-\beta\,\mathbb{D}_{\rm KL}[\pi_\theta,\pi_{ref}]
\Big].
\end{aligned}
\]
where $\mathbb{D}_{\rm KL}[\pi_\theta,\pi_{ref}]$ is the KL divergence between the policy and the reference for completion $i$ and token $t$, $\pi_{ref}$ was the SFT  checkpoint with the best validation RaTEScore, and

\[
\rho_{i,t}(\theta)
=
\frac{
\pi_\theta\!\big(o_{i,t} \mid q, o_{i,<t}\big)
}{
\pi_{\theta_{\mathrm{old}}}\!\big(o_{i,t} \mid q, o_{i,<t}\big)
}, 
\]
\[
\tilde{\rho}_{i,t}(\theta)=
\mathrm{clip}\!\big(
\rho_{i,t}(\theta),\,
1-\epsilon,\,
1+\epsilon
\big).
\]
The optimisation proceeds in an inner–outer loop fashion. For each batch, we first sample the group of completions using the current policy and define a frozen behaviour policy $\pi_{\theta_{\mathrm{old}}}$. The sampled trajectories ${o_i}$, masks $m_{i,t}$, and advantages $A_i$ are then held fixed while minimising $\mathcal{L}_{\mathrm{GRPO}}(\theta)$ for multiple gradient steps. During these updates, $\pi_{\theta_{\mathrm{old}}}$ remains unchanged and is used to compute the importance ratios $\rho_{i,t}(\theta)$. We use three optimisation steps per batch, with $G=4$, $\beta=0.04$, and $\epsilon=0.2$. Validation was executed every fifth of an epoch for GRPO.


\begin{table*}[]
    \centering
    \setlength{\tabcolsep}{4.5pt}

\caption{\label{tab:objective}CXR RRG results. Scores are computed using automated metrics between the \textbf{findings} sections of the generated reports and the radiologist reports. \underline{Underlined} scores indicate a statistically significant difference between the top two highest scores ($p\leq0.05$); the statistical tests are described in Subsection \ref{sec:stat_tests_automated_metrics}. Cell colours indicate relative performance within each metric and dataset (darker = better). The validation split for CXRMate-2 included 991 studies. Only studies with both a findings and an impression section were included in the validation and test sets. RS=RaTEScore, G=GREEN, RE-BS=RadEval BERTScore, RG-XL=RadGraph-XL, CX=CheXbert, CB=CXR-BERT, BS=BERTScore, B4=BLEU, R-L=ROUGE-L.}

\begin{tabular}{lcccccccccc}
\toprule
Model & \#train & RS & G & RE-BS & RG-XL & CX & CB & BS & B4 & R-L\\
\multicolumn{11}{c}{\cellcolor[RGB]{200,200,200} \textit{\textbf{MIMIC-CXR} $n=1\,624$}} \\
EMNLI \citep{miura_improving_2021}&$152\,173$&\cellcolor[RGB]{200,229,236}59.2&\cellcolor[RGB]{184,171,186}39.1&\cellcolor[RGB]{230,238,241}28.6&\cellcolor[RGB]{200,190,202}28.5&\cellcolor[RGB]{227,238,241}31.1&\cellcolor[RGB]{215,209,217}68.1&\cellcolor[RGB]{218,235,239}23.1&\cellcolor[RGB]{239,237,240}6.4&\cellcolor[RGB]{216,234,239}28.0\\
CXRMate \citep{nicolson_longitudinal_2024}&$125\,395$&\cellcolor[RGB]{227,238,241}56.7&\cellcolor[RGB]{213,206,215}36.3&\cellcolor[RGB]{228,238,241}29.1&\cellcolor[RGB]{227,222,227}24.4&\cellcolor[RGB]{197,228,236}35.4&\cellcolor[RGB]{200,190,202}70.2&\cellcolor[RGB]{192,227,235}27.7&\cellcolor[RGB]{229,226,230}9.2&\cellcolor[RGB]{223,236,240}27.0\\
CXRMate-RRG24 \citep{nicolson_e-health_2024}&$550\,395$&\cellcolor[RGB]{214,234,239}58.1&\cellcolor[RGB]{218,211,219}35.8&\cellcolor[RGB]{235,240,242}27.2&\cellcolor[RGB]{207,198,208}27.7&\cellcolor[RGB]{200,229,236}35.1&\cellcolor[RGB]{233,230,234}64.9&\cellcolor[RGB]{204,230,237}25.8&\cellcolor[RGB]{236,233,237}7.6&\cellcolor[RGB]{228,238,241}26.0\\
MAIRA-2 \citep{bannur_maira-2_2024}&$501\,825$&\cellcolor[RGB]{200,229,236}59.2&\cellcolor[RGB]{190,178,192}38.6&\cellcolor[RGB]{153,215,229}37.2&\cellcolor[RGB]{172,157,175}31.6&\cellcolor[RGB]{143,212,228}40.2&\cellcolor[RGB]{196,186,198}70.6&\cellcolor[RGB]{238,241,242}13.9&\cellcolor[RGB]{165,148,167}17.1&\cellcolor[RGB]{144,212,228}34.5\\
Libra \citep{zhang_libra_2025}&$1\,213\,097$&\cellcolor[RGB]{238,241,242}53.7&\cellcolor[RGB]{241,240,242}29.8&\cellcolor[RGB]{238,241,242}24.9&\cellcolor[RGB]{241,240,242}18.1&\cellcolor[RGB]{193,227,235}35.9&\cellcolor[RGB]{182,169,185}72.1&\cellcolor[RGB]{227,238,241}20.5&\cellcolor[RGB]{241,240,242}4.0&\cellcolor[RGB]{238,241,242}21.8\\
CXRMate-ED \citep{nicolson_impact_2025}&$76\,398$&\cellcolor[RGB]{191,226,235}59.9&\cellcolor[RGB]{202,193,204}37.5&\cellcolor[RGB]{219,235,239}30.7&\cellcolor[RGB]{200,190,202}28.5&\cellcolor[RGB]{197,228,236}35.4&\cellcolor[RGB]{101,72,105}\textcolor{white}{78.4}&\cellcolor[RGB]{132,209,226}34.8&\cellcolor[RGB]{221,216,222}10.7&\cellcolor[RGB]{201,230,237}29.8\\
MedGemma (4B) \citep{sellergren_medgemma_2025}&$231\,483$&\cellcolor[RGB]{222,236,240}57.3&\cellcolor[RGB]{231,227,232}33.8&\cellcolor[RGB]{238,241,242}25.7&\cellcolor[RGB]{237,234,237}21.7&\cellcolor[RGB]{228,238,241}30.9&\cellcolor[RGB]{239,237,240}62.8&\cellcolor[RGB]{207,231,238}25.3&\cellcolor[RGB]{240,238,240}6.2&\cellcolor[RGB]{236,240,242}23.9\\
MLRG \citep{liu_enhanced_2025}&-&\cellcolor[RGB]{223,236,240}57.2&\cellcolor[RGB]{222,217,223}35.2&\cellcolor[RGB]{225,237,240}29.7&\cellcolor[RGB]{224,219,225}24.8&\cellcolor[RGB]{224,237,240}31.7&\cellcolor[RGB]{213,205,214}68.6&\cellcolor[RGB]{207,231,238}25.3&\cellcolor[RGB]{236,233,237}7.5&\cellcolor[RGB]{209,232,238}28.9\\
MedVersa \citep{zhou_medversa_2026}&$29\,000\,000$&\cellcolor[RGB]{179,223,233}60.6&\cellcolor[RGB]{188,175,190}38.8&\cellcolor[RGB]{198,229,236}33.3&\cellcolor[RGB]{176,161,178}31.2&\cellcolor[RGB]{190,226,235}36.2&\cellcolor[RGB]{215,208,216}68.3&\cellcolor[RGB]{201,230,237}26.2&\cellcolor[RGB]{177,163,180}16.0&\cellcolor[RGB]{143,212,228}34.6\\
PriorRG \citep{liu_priorrg_2026}&$239\,998$&\cellcolor[RGB]{217,234,239}57.8&\cellcolor[RGB]{220,214,221}35.5&\cellcolor[RGB]{210,232,238}32.0&\cellcolor[RGB]{210,202,211}27.3&\cellcolor[RGB]{210,232,238}33.8&\cellcolor[RGB]{215,209,217}68.1&\cellcolor[RGB]{173,221,232}30.3&\cellcolor[RGB]{208,200,210}12.6&\cellcolor[RGB]{194,227,235}30.5\\
DeepMedix-R1 \citep{lin_toward_2026}&$351\,488$&\cellcolor[RGB]{228,238,241}56.5&\cellcolor[RGB]{235,232,235}33.0&\cellcolor[RGB]{238,241,242}25.8&\cellcolor[RGB]{231,227,232}23.4&\cellcolor[RGB]{238,241,242}26.4&\cellcolor[RGB]{241,240,242}60.4&\cellcolor[RGB]{195,228,236}27.3&\cellcolor[RGB]{237,234,237}7.2&\cellcolor[RGB]{230,238,241}25.6\\
MedGemma 1.5 (4B) \citep{sellergren_medgemma_2026}&-&\cellcolor[RGB]{104,200,222}64.1&\cellcolor[RGB]{144,124,147}41.9&\cellcolor[RGB]{198,229,236}33.3&\cellcolor[RGB]{163,146,166}32.4&\cellcolor[RGB]{25,176,209}47.1&\cellcolor[RGB]{219,213,220}67.6&\cellcolor[RGB]{137,210,227}34.3&\cellcolor[RGB]{185,172,187}15.3&\cellcolor[RGB]{156,216,230}33.7\\
\textbf{CXRMate-2} (ours)&$313\,503$&\cellcolor[RGB]{0,169,206}\textbf{\underline{67.6}}&\cellcolor[RGB]{54,16,60}\textbf{\textcolor{white}{\underline{46.6}}}&\cellcolor[RGB]{0,169,206}\textbf{\underline{45.5}}&\cellcolor[RGB]{54,16,60}\textbf{\textcolor{white}{\underline{40.3}}}&\cellcolor[RGB]{0,169,206}\textbf{48.3}&\cellcolor[RGB]{54,16,60}\textbf{\textcolor{white}{81.2}}&\cellcolor[RGB]{0,169,206}\textbf{\underline{45.3}}&\cellcolor[RGB]{54,16,60}\textbf{\textcolor{white}{\underline{24.5}}}&\cellcolor[RGB]{0,169,206}\textbf{\underline{42.1}}\\

\multicolumn{11}{c}{\cellcolor[RGB]{200,200,200} \textit{\textbf{CheXpert Plus} $n=62$}} \\
MAIRA-2 \citep{bannur_maira-2_2024}&$501\,825$&\cellcolor[RGB]{238,235,238}51.1&\cellcolor[RGB]{186,225,234}27.2&\cellcolor[RGB]{215,208,216}25.2&\cellcolor[RGB]{229,238,241}19.6&\cellcolor[RGB]{224,219,225}28.4&\cellcolor[RGB]{113,203,223}71.8&\cellcolor[RGB]{76,42,81}\textcolor{white}{19.9}&\cellcolor[RGB]{225,237,240}4.8&\cellcolor[RGB]{224,218,225}22.6\\
Libra \citep{zhang_libra_2025}&$1\,213\,097$&\cellcolor[RGB]{241,240,242}49.6&\cellcolor[RGB]{238,241,242}22.4&\cellcolor[RGB]{218,212,220}24.8&\cellcolor[RGB]{238,241,242}17.1&\cellcolor[RGB]{226,221,227}28.3&\cellcolor[RGB]{72,191,217}74.9&\cellcolor[RGB]{146,125,149}15.5&\cellcolor[RGB]{238,241,242}2.5&\cellcolor[RGB]{241,240,242}18.4\\
MedGemma (4B) \citep{sellergren_medgemma_2025}&$231\,483$&\cellcolor[RGB]{234,231,235}51.7&\cellcolor[RGB]{236,240,242}23.5&\cellcolor[RGB]{236,233,237}22.2&\cellcolor[RGB]{238,241,242}17.4&\cellcolor[RGB]{241,240,242}25.4&\cellcolor[RGB]{226,237,241}57.8&\cellcolor[RGB]{213,206,215}9.2&\cellcolor[RGB]{236,240,242}3.4&\cellcolor[RGB]{239,237,240}19.9\\
MedVersa \citep{zhou_medversa_2026}&$29\,000\,000$&\cellcolor[RGB]{241,240,242}49.4&\cellcolor[RGB]{237,240,242}23.2&\cellcolor[RGB]{241,240,242}20.4&\cellcolor[RGB]{238,241,242}17.6&\cellcolor[RGB]{236,233,237}27.1&\cellcolor[RGB]{238,241,242}51.5&\cellcolor[RGB]{241,240,242}1.6&\cellcolor[RGB]{236,240,242}3.5&\cellcolor[RGB]{238,236,239}20.1\\
DeepMedix-R1 \citep{lin_toward_2026}&$351\,488$&\cellcolor[RGB]{235,233,236}51.5&\cellcolor[RGB]{198,229,236}26.6&\cellcolor[RGB]{239,237,240}21.4&\cellcolor[RGB]{224,236,240}20.4&\cellcolor[RGB]{226,221,227}28.3&\cellcolor[RGB]{207,231,238}61.6&\cellcolor[RGB]{206,197,207}10.1&\cellcolor[RGB]{231,239,241}4.2&\cellcolor[RGB]{229,226,230}21.8\\
MedGemma 1.5 (4B) \citep{sellergren_medgemma_2026}&-&\cellcolor[RGB]{232,229,233}51.9&\cellcolor[RGB]{219,235,239}25.3&\cellcolor[RGB]{241,240,242}19.7&\cellcolor[RGB]{238,241,242}17.7&\cellcolor[RGB]{227,222,227}28.2&\cellcolor[RGB]{217,234,239}59.9&\cellcolor[RGB]{210,202,211}9.6&\cellcolor[RGB]{237,240,242}3.2&\cellcolor[RGB]{232,228,232}21.5\\
\textbf{CXRMate-2} (ours)&$313\,503$&\cellcolor[RGB]{54,16,60}\textbf{\textcolor{white}{\underline{60.8}}}&\cellcolor[RGB]{0,169,206}\textbf{32.7}&\cellcolor[RGB]{54,16,60}\textbf{\textcolor{white}{\underline{34.3}}}&\cellcolor[RGB]{0,169,206}\textbf{\underline{30.4}}&\cellcolor[RGB]{54,16,60}\textbf{\textcolor{white}{35.3}}&\cellcolor[RGB]{0,169,206}\textbf{79.6}&\cellcolor[RGB]{54,16,60}\textbf{\textcolor{white}{21.1}}&\cellcolor[RGB]{0,169,206}\textbf{\underline{12.5}}&\cellcolor[RGB]{54,16,60}\textbf{\textcolor{white}{\underline{31.9}}}\\

\multicolumn{11}{c}{\cellcolor[RGB]{200,200,200} \textit{\textbf{ReXgradient} $n=10\,000$}} \\
Libra \citep{zhang_libra_2025}&$1\,213\,097$&\cellcolor[RGB]{238,241,242}56.2&\cellcolor[RGB]{240,238,240}44.8&\cellcolor[RGB]{238,241,242}20.3&\cellcolor[RGB]{241,240,242}18.5&\cellcolor[RGB]{238,241,242}6.1&\cellcolor[RGB]{241,240,242}31.9&\cellcolor[RGB]{238,241,242}23.8&\cellcolor[RGB]{241,240,242}4.6&\cellcolor[RGB]{238,241,242}21.9\\
MedGemma (4B) \citep{sellergren_medgemma_2025}&$231\,483$&\cellcolor[RGB]{198,229,236}61.2&\cellcolor[RGB]{63,27,69}\textcolor{white}{57.7}&\cellcolor[RGB]{229,238,241}25.1&\cellcolor[RGB]{221,215,222}26.2&\cellcolor[RGB]{150,214,229}29.7&\cellcolor[RGB]{160,143,163}62.4&\cellcolor[RGB]{221,236,240}30.2&\cellcolor[RGB]{241,240,242}5.3&\cellcolor[RGB]{235,240,242}24.5\\
MedVersa \citep{zhou_medversa_2026}&$29\,000\,000$&\cellcolor[RGB]{238,241,242}55.9&\cellcolor[RGB]{241,240,242}43.4&\cellcolor[RGB]{238,241,242}21.7&\cellcolor[RGB]{239,237,240}21.4&\cellcolor[RGB]{228,238,241}13.8&\cellcolor[RGB]{241,240,242}30.6&\cellcolor[RGB]{237,240,242}25.4&\cellcolor[RGB]{241,240,242}4.4&\cellcolor[RGB]{236,240,242}24.4\\
MedGemma 1.5 (4B) \citep{sellergren_medgemma_2026}&-&\cellcolor[RGB]{140,211,227}64.2&\cellcolor[RGB]{73,39,79}\textcolor{white}{57.3}&\cellcolor[RGB]{209,232,238}29.2&\cellcolor[RGB]{157,139,160}34.1&\cellcolor[RGB]{54,185,214}40.2&\cellcolor[RGB]{155,137,158}63.4&\cellcolor[RGB]{188,226,235}34.8&\cellcolor[RGB]{228,224,229}10.8&\cellcolor[RGB]{181,223,233}33.3\\
\textbf{CXRMate-2} (ours)&$313\,503$&\cellcolor[RGB]{0,169,206}\textbf{\underline{68.9}}&\cellcolor[RGB]{54,16,60}\textbf{\textcolor{white}{58.1}}&\cellcolor[RGB]{0,169,206}\textbf{\underline{45.6}}&\cellcolor[RGB]{54,16,60}\textbf{\textcolor{white}{\underline{41.7}}}&\cellcolor[RGB]{0,169,206}\textbf{44.9}&\cellcolor[RGB]{54,16,60}\textbf{\textcolor{white}{\underline{79.0}}}&\cellcolor[RGB]{0,169,206}\textbf{\underline{47.9}}&\cellcolor[RGB]{54,16,60}\textbf{\textcolor{white}{\underline{28.1}}}&\cellcolor[RGB]{0,169,206}\textbf{\underline{45.2}}\\

    \bottomrule
    \end{tabular}
\end{table*}

\subsubsection{Automated metrics \& their applicability}


We evaluate CXR RRG using a diverse set of automated metrics, grouped as follows:
\begin{itemize}
    \item \textbf{Lexical overlap:}
    BLEU (\textbf{B4}) \citep{papineni_bleu_2002},
    ROUGE-L (\textbf{R-L}) \citep{lin_automatic_2004}.
    
    \item \textbf{General-domain semantic similarity:}
    BERTScore (\textbf{BS}) \citep{zhang_bertscore_2019}.
    
    \item \textbf{Radiology report semantic similarity:}
    CXR-BERT (\textbf{CB}) \citep{avidan_making_2022},
    RadEval BERTScore (\textbf{RE-BS}) \citep{xu_radeval_2025}.
    
    \item \textbf{Entity-level semantic alignment:}
    RaTEScore (\textbf{RS}) \citep{zhao_ratescore_2024}.
    
    \item \textbf{Multi-label findings classification:}
    CheXbert (\textbf{CX}) \citep{smit_combining_2020}.
    
    \item \textbf{Graph-level factual correctness:}
    RadGraph-XL (\textbf{RG-XL}).
    
    \item \textbf{LLM-based clinical error assessment:}
    GREEN (\textbf{G}) \citep{ostmeier_green_2024}.
\end{itemize}
For CheXbert, the macro-averaged F1 was computed between the 14 CheXbert findings extracted from the generated and radiologist reports. ``No mention'', ``negative'', and ``uncertain'' were considered negative, while ``positive'' was considered positive.

The literature suggests that these metrics exhibit only partial alignment with radiologist raters. \cite{yu_evaluating_2023} report that GREEN, ROUGE-L, BERTScore, and BLEU obtain Kendall's $\tau$ values of 0.63, 0.56, 0.49, and 0.35 on the ReXVal benchmark, respectively. Similarly, \cite{zhao_ratescore_2024} report Kendall's $\tau$ values of 0.46 and 0.29 for RaTEScore and CheXbert on the RaTE-Eval benchmark, respectively. This indicates only moderate correlation with radiologist raters and motivates qualitative evaluation to more reliably assess CXR RRG.

\subsubsection{Significance testing} \label{sec:stat_tests_automated_metrics}


We also perform statistical testing to determine whether
there is a significant difference in metric scores between
models. As all models were evaluated on the same test set,
per-study metric scores were treated as paired observations.
A one-way ANOVA was used to assess whether metric
scores differed between models. When the ANOVA result
was significant, Tukey’s HSD test was conducted for post-
hoc pairwise comparisons. As CheXbert was evaluated at the dataset level without per-study metric scores, it is not suitable for paired comparison.

\subsection{Results \& discussion} \label{sec:results_cxrmate2}

Results comparing CXRMate-2 to benchmark models in the literature are presented in Table \ref{tab:objective}. The benchmark models are described in Appendix \ref{sec:benchmark_models}. CXRMate-2 achieves SOTA performance across all metrics and datasets, attaining statistically significant improvements over the next best model in a majority of cases. These results indicate that our earlier qualitative retrospective evaluation was conducted using a SOTA CXR RRG model with leading benchmark performance.



\subsubsection{Model architecture} \label{sec:development}

\paragraph{\textbf{Baseline:}}
The purpose of this experiment was to establish a controlled baseline for subsequent model improvements. To this end, we reproduced MAIRA-2 by training exclusively on MIMIC-CXR (a full replication is impossible, as we do not have access to the private data used to train MAIRA-2). The adapter was randomly initialised, the decoder was initialised from Vicuna 7B v1.5, and the RAD-DINO visual encoder was kept frozen. 

This model, denoted as Reproduction in Table \ref{tab:reproduction}, achieved performance on the MIMIC-CXR test set comparable to MAIRA-2 despite the substantial difference in training data scale and diversity. This likely reflects the close alignment between the training and evaluation distributions, as the reproduction model was trained solely on MIMIC-CXR. This baseline therefore provides a controlled reference point for isolating the impact of subsequent architectural and training improvements.

\begin{table}[h]
\centering
\setlength{\tabcolsep}{4.5pt}
\caption{Reduction in self-attention computational complexity relative to \textit{Reproduction}. $^\ddagger$Average prompt length over the MIMIC-CXR test set ($n=1\,624$; studies with both a findings and an impression section were considered).}
\label{tab:attention_reduction}
\rowcolors{2}{gray!15}{white}
\begin{tabular}{p{3.1cm}p{1.25cm}p{1.25cm}p{1.25cm}}
\toprule
Setting & Average prompt length & Relative complexity & Reduction \\
\midrule
Reproduction & 6407$^\ddagger$ & 1.0 & 0.0\% \\
Q-Adapter & 3586$^\ddagger$ & 0.313 & 68.7\% \\
Structured multimodal temporal embeddings & 607$^\ddagger$ & 0.009 & 99.1\% \\
\bottomrule
\end{tabular}
\end{table}

\paragraph{\textbf{Q-Adapter:}}
Next, the Q-Adapter is added to the reproduced model. This leads to a slight decrease in performance to the RRG metrics and an increase to the lexical metrics. This is likely due to the compression of the visual features from $1\,369$ inputs per CXR to 128. As shown in Table \ref{tab:attention_reduction}, this reduces the average prompt length from $6\,407$ inputs to $3\,586$, reducing attention-related activation memory (which scales quadratically with sequence length) by approximately 68.7\% per attention head in the decoder (assuming a report of 128 inputs). This trade-off prioritises reduced self-attention computational complexity and enables tractable training with resource-intensive RL.

\paragraph{\textbf{Structured multimodal temporal embeddings:}}
Switching to the structured multimodal temporal embeddings yielded the best performance across all metrics. This improvement could be attributable to the richer input representation over MAIRA-2. CXRMate-2 includes all CXRs from the study and its prior (whereas MAIRA-2 includes one frontal and one lateral from the study and one frontal from the prior). It also additionally includes the history section, and time-delta embeddings. Explicitly signalling each input type to the decoder via source embeddings, rather than through instructions, may be beneficial because it provides a more structured conditioning signal and may mitigate attention dilution effects, as discussed in \cite{nicolson_impact_2025}. As shown in Table \ref{tab:attention_reduction}, this approach also substantially reduced self-attention computational complexity, with the average prompt length reduced further from $3\,586$ to 607. Combined with the Q-Adapter, this reduces attention-related activation memory by approximately 99.1\% per attention head in the decoder (assuming a report of 128 inputs).

By reducing self-attention computational complexity, the Q-Adapter and structured multimodal temporal embeddings make GRPO tractable for CXR RRG under modest compute resources. Without these modifications, applying GRPO---which requires a reference model and multiple completions per prompt---would be infeasible in this setting. This substantially lowers both implementation overhead and compute requirements, making GRPO practical where it would otherwise be prohibitive.

\begin{table}[]
    \centering
    \setlength{\tabcolsep}{3.75pt}
    \small

\caption{\label{tab:reproduction}Merging  and improving CXRMate-ED and MAIRA-2. + and \bentarrow~indicate an addition or modification to the model in the prior row, respectively. Following MAIRA-2, only studies with a findings section and a frontal view were considered for training and testing, and each model was optimised to generate the findings section. Using only studies from MIMIC-CXR, the validation and test splits included $1\,151$ and $2\,210$ studies, respectively. $^\ddagger$Inputs from MIMIC-IV-ED were not considered \citep{johnson_mimic-iv-ed_2023, johnson_mimic-iv_2023}. Cell colours indicate relative performance within each metric (darker = better).}
\begin{tabular}{p{3.3cm}ccccc}
\toprule
Model & \#train & RS & G & BS & B4\\
\midrule
MAIRA-2&$501\,825$&\cellcolor[RGB]{238,241,242}55.8&\cellcolor[RGB]{241,240,242}37.1&\cellcolor[RGB]{127,207,225}35.7&\cellcolor[RGB]{222,217,223}12.8\\
Reproduction&$146\,138$&\cellcolor[RGB]{232,239,241}56.1&\cellcolor[RGB]{221,215,222}38.0&\cellcolor[RGB]{238,241,242}28.7&\cellcolor[RGB]{241,240,242}11.7\\
\bentarrow Q-Adapter&$146\,138$&\cellcolor[RGB]{238,241,242}55.8&\cellcolor[RGB]{239,237,240}37.4&\cellcolor[RGB]{227,238,241}30.9&\cellcolor[RGB]{210,202,211}13.1\\
\bentarrow Structured multimodal temporal embeddings$^\ddagger$&$146\,138$&\cellcolor[RGB]{0,169,206}\textbf{57.7}&\cellcolor[RGB]{54,16,60}\textbf{\textcolor{white}{39.8}}&\cellcolor[RGB]{0,169,206}\textbf{39.0}&\cellcolor[RGB]{54,16,60}\textbf{\textcolor{white}{15.1}}\\

    \bottomrule
    \end{tabular}
\end{table}

\subsubsection{Language model}

We investigated alternatives in Table \ref{tab:language_model} to the Vicuna 7B v1.5 LLM used as the decoder for MAIRA-2. Specifically, we evaluated two families of LLMs: LLaMA 3 and Qwen 2.5, each at multiple model sizes. These were all pre-trained LLMs and not their instruction-tuned variants. Overall, the 3B Llama 3.2 LLM performed best, providing a moderate boost in performance across all metrics.

\begin{table}[]
    \centering
    \setlength{\tabcolsep}{4.95pt}
    \small

\caption{\label{tab:language_model}Decoder selection. Following MAIRA-2, only studies with a findings section and a frontal view were considered for training and testing, and each model was optimised to generate the findings section. Using only studies from MIMIC-CXR, the training, validation, and test splits included $146\,138$, $1\,151$, and $2\,210$ studies, respectively. Cell colours indicate relative performance within each metric (darker = better).}
\begin{tabular}{lcccc}
\toprule
LLM decoder & RS & G & BS & B4\\
\midrule
Vicuna 7B v1.5&\cellcolor[RGB]{174,159,176}57.7&\cellcolor[RGB]{148,213,228}39.8&\cellcolor[RGB]{201,191,202}39.0&\cellcolor[RGB]{179,223,233}15.1\\
LLaMA 3.2 1B&\cellcolor[RGB]{121,96,125}\textcolor{white}{58.1}&\cellcolor[RGB]{128,208,225}40.0&\cellcolor[RGB]{121,96,125}\textcolor{white}{40.5}&\cellcolor[RGB]{11,172,207}17.0\\
LLaMA 3.2 3B&\cellcolor[RGB]{54,16,60}\textbf{\textcolor{white}{58.5}}&\cellcolor[RGB]{0,169,206}\textbf{41.0}&\cellcolor[RGB]{54,16,60}\textbf{\textcolor{white}{41.4}}&\cellcolor[RGB]{0,169,206}\textbf{17.1}\\
LLaMA 3.1 8B&\cellcolor[RGB]{230,226,231}57.0&\cellcolor[RGB]{208,232,238}39.0&\cellcolor[RGB]{146,126,150}40.1&\cellcolor[RGB]{104,200,222}16.1\\
Qwen 2.5 0.5B&\cellcolor[RGB]{241,240,242}56.5&\cellcolor[RGB]{238,241,242}37.9&\cellcolor[RGB]{241,240,242}36.9&\cellcolor[RGB]{238,241,242}13.1\\
Qwen 2.5 1.5B&\cellcolor[RGB]{195,184,197}57.5&\cellcolor[RGB]{196,228,236}39.2&\cellcolor[RGB]{218,211,219}38.5&\cellcolor[RGB]{200,229,236}14.7\\
Qwen 2.5 3B&\cellcolor[RGB]{241,240,242}56.5&\cellcolor[RGB]{213,233,239}38.9&\cellcolor[RGB]{238,236,239}37.5&\cellcolor[RGB]{236,240,242}13.5\\
Qwen 2.5 7B&\cellcolor[RGB]{238,235,238}56.8&\cellcolor[RGB]{225,237,240}38.6&\cellcolor[RGB]{152,133,155}40.0&\cellcolor[RGB]{104,200,222}16.1\\

        \bottomrule
    \end{tabular}
\end{table}

\subsubsection{Reinforcement learning (RL)} \label{sec:rl_results}

We examined improvements to the RL training stage (Table~\ref{tab:reinforcement_learning}). Following CXRMate-ED, we first applied SCST with a composite reward comprising CXR-BERT, BERTScore, and ARN. LoRA was applied to the query and value projection matrices in each decoder attention layer, using rank $r=16$, scaling factor $\alpha=32$, and a dropout rate of 0.05 during training \citep{hu_lora_2022}.

This yielded modest gains over SFT across all metrics. Replacing SCST with GRPO while retaining the same reward produced further consistent improvements, indicating more stable and effective policy optimisation. This application of GRPO is enabled by the Q-Adapter and structured multimodal temporal embeddings, which reduce self-attention computational complexity; without these modifications, GRPO---requiring a reference model and multiple completions per prompt---would be computationally infeasible under modest compute resources.

Incorporating RaTEScore into the composite reward led to additional gains. This improvement is likely attributable to RaTEScore’s stronger alignment with radiologist ratings, as evidenced by its Kendall’s $\tau$ of 0.46 on the RaTE-Eval benchmark, where it outperforms other metrics \citep{zhao_ratescore_2024}.

Finally, training the full decoder rather than restricting updates to LoRA layers resulted in the largest overall improvement, achieving the best performance across all metrics. These results demonstrate that both improved policy optimisation and reward design contribute to performance gains, with full decoder fine-tuning providing complementary benefits beyond parameter-efficient fine-tuning.

\begin{table}[]
    \centering
    \setlength{\tabcolsep}{4.95pt}
    \small

\caption{\label{tab:reinforcement_learning}RL improvements. $^\ddagger$The CXRMate-ED reward was used, which included CXR-BERT, BERTScore, and ARN, with a weight of: 0.45, 0.45, and 0.1, respectively. $^\S$Adds RaTEScore to the CXRMate-ED reward, with RaTEScore, CXR-BERT, BERTScore each having a weight of 0.3, and ARN with a weight of 0.1. + and \bentarrow~indicate an addition or modification to the model in the prior row, respectively. Following MAIRA-2, only studies with a findings section and a frontal view were considered for training and testing, and each model was optimised to generate the findings section. Using only studies from MIMIC-CXR, the training, validation, and test splits included $146\,138$, $1\,151$, and $2\,210$ studies, respectively. LoRA was applied to the query and value projection matrices in each decoder attention layer, using rank $r=16$, scaling factor $\alpha=32$, and a dropout rate of 0.05 during training \citep{hu_lora_2022}. Cell colours indicate relative performance within each metric (darker = better).}
\begin{tabular}{p{2.75cm}cccc}
\toprule
Fine-tuning & RS & G & BS & B4\\
\midrule
SFT&\cellcolor[RGB]{241,240,242}58.3&\cellcolor[RGB]{238,241,242}41.1&\cellcolor[RGB]{241,240,242}41.0&\cellcolor[RGB]{238,241,242}16.8\\
+ RL SCST LoRA$^\ddagger$&\cellcolor[RGB]{236,233,237}58.7&\cellcolor[RGB]{228,238,241}41.5&\cellcolor[RGB]{212,204,213}42.0&\cellcolor[RGB]{200,229,236}17.4\\
\bentarrow RL GRPO LoRA$^\ddagger$&\cellcolor[RGB]{233,230,234}58.8&\cellcolor[RGB]{216,234,239}41.7&\cellcolor[RGB]{199,189,200}42.2&\cellcolor[RGB]{200,229,236}17.4\\
\bentarrow RL GRPO LoRA$^\S$&\cellcolor[RGB]{209,201,210}59.3&\cellcolor[RGB]{152,215,229}42.3&\cellcolor[RGB]{174,159,176}42.5&\cellcolor[RGB]{132,209,226}17.8\\
\bentarrow RL GRPO$^\S$&\cellcolor[RGB]{54,16,60}\textbf{\textcolor{white}{60.7}}&\cellcolor[RGB]{0,169,206}\textbf{43.1}&\cellcolor[RGB]{54,16,60}\textbf{\textcolor{white}{43.5}}&\cellcolor[RGB]{0,169,206}\textbf{18.3}\\

        \bottomrule
    \end{tabular}
\end{table}

\subsubsection{Datasets} \label{sec:datasets}

We next evaluated the impact of the training data (Table~\ref{tab:datasets}). Models trained on the same dataset as the test set achieved the strongest performance, indicating substantial domain specificity. For example, training on the ReXgradient training set yielded the best RS and B4 on the ReXgradient test set. 

Cross-dataset generalisation was limited: models trained on CheXpert Plus or ReXgradient showed marked degradation when evaluated on MIMIC-CXR, particularly on the lexical metrics. This likely reflects institutional differences in report structure, phrasing, and terminology.

Training on all datasets consistently improved robustness across domains, achieving the best scores across all metrics on CheXpert Plus and competitive results on both MIMIC-CXR and ReXgradient. Notably, CheXpert Plus has substantially fewer training examples, suggesting that incorporating data from other institutions can meaningfully improve performance when local data are limited.

\begin{table}[]
    \centering
    \setlength{\tabcolsep}{4.95pt}
    \small

\caption{\label{tab:datasets}Dataset impact. Evaluation was performed on the \textbf{findings} section of each test set. Only studies with a findings and impression section were considered for training and testing. Each model was optimised to generate both the findings and impression sections. Cell colours indicate relative performance within each metric and dataset (darker = better). The validation set included 991 studies.}
\begin{tabular}{lccccc}
\toprule
Training dataset & \#train & RS & G & BS & B4\\
\midrule
\multicolumn{6}{c}{\cellcolor[RGB]{200,200,200} \textit{\textbf{MIMIC-CXR} $n=1\,624$}} \\
MIMIC-CXR&$125\,417$&\cellcolor[RGB]{0,169,206}\textbf{65.4}&\cellcolor[RGB]{73,39,79}\textcolor{white}{44.6}&\cellcolor[RGB]{12,172,207}43.7&\cellcolor[RGB]{54,16,60}\textbf{\textcolor{white}{23.1}}\\
CheXpert Plus&$48\,086$&\cellcolor[RGB]{238,241,242}53.1&\cellcolor[RGB]{241,240,242}27.7&\cellcolor[RGB]{238,241,242}7.3&\cellcolor[RGB]{241,240,242}4.0\\
ReXgradient&$140\,000$&\cellcolor[RGB]{238,241,242}53.3&\cellcolor[RGB]{240,238,240}29.5&\cellcolor[RGB]{216,234,239}18.7&\cellcolor[RGB]{241,240,242}4.2\\
All&$313\,503$&\cellcolor[RGB]{98,198,221}62.5&\cellcolor[RGB]{54,16,60}\textbf{\textcolor{white}{45.6}}&\cellcolor[RGB]{0,169,206}\textbf{44.7}&\cellcolor[RGB]{121,95,125}\textcolor{white}{19.3}\\

\multicolumn{6}{c}{\cellcolor[RGB]{200,200,200} \textit{\textbf{CheXpert Plus} $n=62$}} \\
MIMIC-CXR&$125\,417$&\cellcolor[RGB]{231,227,232}48.7&\cellcolor[RGB]{238,241,242}23.6&\cellcolor[RGB]{210,202,211}9.0&\cellcolor[RGB]{238,241,242}2.5\\
CheXpert Plus&$48\,086$&\cellcolor[RGB]{140,118,143}\textcolor{white}{52.3}&\cellcolor[RGB]{238,241,242}23.5&\cellcolor[RGB]{105,77,110}\textcolor{white}{18.7}&\cellcolor[RGB]{154,215,229}6.1\\
ReXgradient&$140\,000$&\cellcolor[RGB]{241,240,242}47.0&\cellcolor[RGB]{238,241,242}23.4&\cellcolor[RGB]{241,240,242}-0.1&\cellcolor[RGB]{238,241,242}2.8\\
All&$313\,503$&\cellcolor[RGB]{54,16,60}\textbf{\textcolor{white}{54.2}}&\cellcolor[RGB]{0,169,206}\textbf{30.2}&\cellcolor[RGB]{54,16,60}\textbf{\textcolor{white}{22.0}}&\cellcolor[RGB]{0,169,206}\textbf{8.6}\\

\multicolumn{6}{c}{\cellcolor[RGB]{200,200,200} \textit{\textbf{ReXgradient} $n=10\,000$}} \\
MIMIC-CXR&$125\,417$&\cellcolor[RGB]{200,229,236}58.1&\cellcolor[RGB]{217,211,218}48.9&\cellcolor[RGB]{200,229,236}28.4&\cellcolor[RGB]{241,239,241}4.9\\
CheXpert Plus&$48\,086$&\cellcolor[RGB]{238,241,242}52.7&\cellcolor[RGB]{241,240,242}44.9&\cellcolor[RGB]{238,241,242}16.6&\cellcolor[RGB]{241,240,242}3.0\\
ReXgradient&$140\,000$&\cellcolor[RGB]{0,169,206}\textbf{66.3}&\cellcolor[RGB]{86,54,91}\textcolor{white}{55.0}&\cellcolor[RGB]{10,172,207}45.6&\cellcolor[RGB]{54,16,60}\textbf{\textcolor{white}{27.3}}\\
All&$313\,503$&\cellcolor[RGB]{52,184,214}64.7&\cellcolor[RGB]{54,16,60}\textbf{\textcolor{white}{56.0}}&\cellcolor[RGB]{0,169,206}\textbf{46.3}&\cellcolor[RGB]{84,52,90}\textcolor{white}{25.2}\\

        \bottomrule
    \end{tabular}
\end{table}

\section{Qualitative retrospective evaluation} \label{sec:qual_section}

\subsection{Methodology} \label{sec:retro_meth}

We conducted a qualitative retrospective evaluation to determine radiologist preferences between radiology reports generated from a SOTA CXR RRG model (CXRMate-2, presented in Section \ref{sec:cxrmate2_section}) and those written by radiologists. A \textit{study} is a single imaging examination for a patient, comprising one or more CXRs acquired during the same session and accompanied by an associated radiology report. For each study, reports were compared under a blinded, randomised presentation by three consultant radiologists (raters). They recorded a pairwise preference for either the generated or radiologist report, with an explicit option to indicate no preference when the reports were judged equivalent. Raters also recorded one or more reasons for their preferences, with the reasons corresponding to precision, recall, and readability.

There is little empirical guidance in the literature regarding statistical power, number of studies, or the number of raters required for such evaluations; therefore, this qualitative retrospective study was designed as a pilot. The analysis was therefore exploratory, aiming to estimate key design parameters---such as effect sizes, sample size requirements, and rater numbers---to inform subsequent, adequately powered qualitative retrospective evaluations.

\begin{figure*}[]
  \centering
  \includegraphics[width=\textwidth]{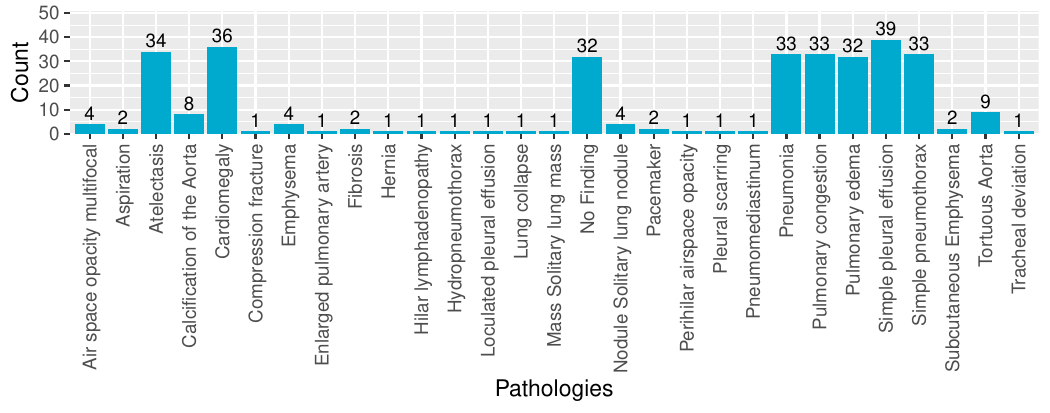}
  \caption{Findings in the 120 selected studies (eight findings were targeted for analysis---those shown in the figure that occur in 30 studies or more). Note that a study can have one or more findings present.}
  \label{fig:pathology_distribution}
\end{figure*}

\begin{figure*}[]
  \centering
  \includegraphics[width=\textwidth]{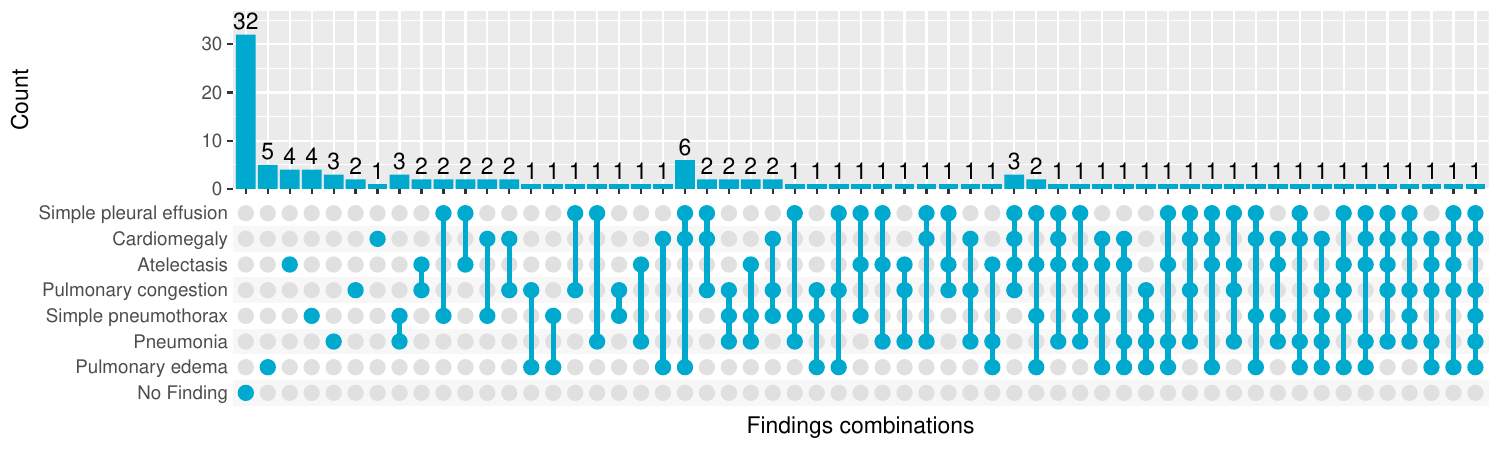}
  \caption{UpSet plot of the 120 selected studies---only the eight targeted findings described in Subsection \ref{sec:selection} are included in the plot. Sorted by degree, then count.}
  \label{fig:upset}
\end{figure*}

\begin{figure}[]
  \centering
  \includegraphics[width=\columnwidth]{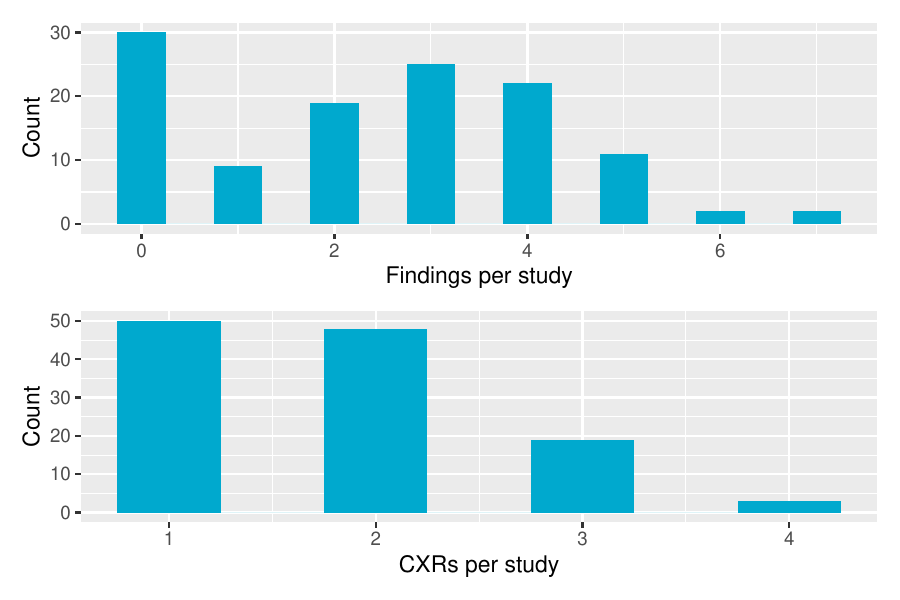}
  \caption{Distributions for the 120 selected studies. \textbf{Top}: Number of findings per study (all findings of SRR-BERT are considered here, unlike in Figure \ref{fig:upset}). Studies with zero findings corresponded to studies labelled `No Finding'. \textbf{Bottom}: Number of CXRs per study. Note, these do not consider the prior studies.}
  \label{fig:per_study_distributions}
\end{figure}

\subsubsection{Data} \label{sec:data}

Studies were selected from the test split of the MIMIC-CXR dataset \citep{johnson_mimic-cxr_2019, johnson_mimic-cxr_2024}. MIMIC-CXR is a publicly-available dataset of CXR studies and free-text radiology reports performed at the Beth Israel Deaconess Medical Center in Boston, Massachusetts, between 2011 and 2016. A total of 120 CXR studies were selected, each corresponding to a unique patient, with the selection procedure described in Subsection \ref{sec:selection}. To limit the workload of the raters to approximately four hours---and assuming an estimated review time of two minutes per study---120 studies were selected. Out of the 120 studies, 107 had a prior study available. All available CXRs for a given study and its prior were presented to the raters. The CXRs were in DICOM format and included anteroposterior, posteroanterior, and lateral projections. From the radiology reports of each study, the indication, history, findings, and impression sections were presented to the raters.

The findings for the selected studies are shown in Figure \ref{fig:pathology_distribution}, where eight common findings were present in a minimum of 32 studies. The findings were extracted from the radiology reports using SRR-BERT \citep{delbrouck_automated_2025}. The UpSet plot in Figure \ref{fig:upset} provides a count of the different combinations of findings. Along with Figure \ref{fig:per_study_distributions} top, it can be seen that a large portion of the selected studies are complex, with 41\% having three or more findings. The number of CXR per study in Figure \ref{fig:per_study_distributions} bottom shows that most studies included one or two CXRs. 

\subsubsection{Raters and ethics}
Three consultant radiologists with 6, 16, and 33 years of post-consultancy (attending-level) experience from the Princess Alexandra Hospital (PAH) in Brisbane, Australia, participated as raters. The study received approval from the Metro South Human Research Ethics Committee (HREC/2025/QMS/116948). Rater participation required written and informed consent, while use of patient data relied on the publicly available, fully de-identified MIMIC-CXR dataset. 

\subsubsection{Study selection}\label{sec:selection}

\begin{figure}[]
  \centering
  \includegraphics[width=\columnwidth]{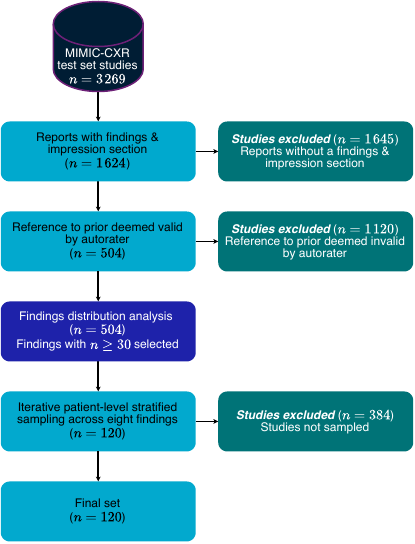}
  \caption{Study selection flow diagram.}
  \label{fig:selection}
\end{figure}

Studies were selected from the MIMIC-CXR test set ($n=3\,269$) according to multiple criteria, as shown in Figure \ref{fig:selection}. Only studies with a radiology report containing both a findings and an impression section were considered, reducing the pool of studies to $1\,624$ (this requirement was not applied to the prior studies). 


A key challenge in study selection was the presence of gaps in patients’ study histories in the MIMIC-CXR dataset. Radiologists often compare a study with a prior study from the same patient. However, in MIMIC-CXR, the specific prior study referenced in the report is sometimes missing. In such cases, we define the available prior as the chronologically closest preceding study present in the dataset. This available prior may not correspond to the true prior referenced in the report, leading to inconsistencies---for example, interval changes may be described for findings not present in the available prior. In addition, reports frequently reference findings from priors of other modalities (e.g., CT), which were not available to CXRMate-2. To resolve this ambiguity, Gemini 2.5 Flash was used as an autorater to determine if a study should be excluded or not, based on its reference to the available prior \citep{comanici_gemini_2025}. The autorater was used to determine:
\begin{itemize}
    \item Whether the available prior corresponds to the true prior referenced in the study report.
    \item Whether a study with no available prior refrains from referencing any prior.
    \item Whether the study refrains from referencing findings from other imaging modalities.
\end{itemize}
A study was excluded if any of these conditions were not met. Figure \ref{fig:prior_study_validation_prompt} shows the header of the autorater prompt; the remainder of the prompt consisted of the radiology reports of the study and its prior, their respective dates and times, and the time difference between them. From an initial pool of $1\,624$ candidate studies, 504 were judged to meet the criteria.

The autorater was not formally evaluated, as it was used only for dataset curation, rather than as part of the model or evaluation pipeline. Any residual errors are therefore expected to introduce noise in study selection rather than systematic bias, since both generated and radiologist reports are evaluated on the same set of studies.

To obtain a balanced sample across a subset of common findings, we first analysed the distribution of findings in the remaining 504 studies (Figure \ref{fig:pathology_distribution_valid}). We targeted findings with at least 30 available studies, namely atelectasis, cardiomegaly, no findings, pneumonia, pulmonary congestion, pulmonary edema, simple pleural effusion, and simple pneumothorax. 

To approximate a balanced distribution across these eight findings, we applied an iterative stratified sampling procedure at the patient level. At each iteration, we first identified the finding with the fewest selected studies. We then randomly sampled (without replacement) a study that was positive for this finding and belonged to a patient not yet included in the sample. After each selection, counts for all findings present in the sampled study were updated. Sampling continued until 120 studies were selected, producing an approximately uniform distribution across the targeted findings while ensuring one study per patient. The final findings distribution is shown in Figure \ref{fig:pathology_distribution}.

\subsubsection{Evaluation procedure}

The evaluation was conducted using a secure, web-based platform capable of displaying the full bit depth of the DICOM images on a diagnostic medical display. Further description of the evaluation platform, including a mock-up of the evaluation platform is described in Section \ref{sec:mockup}. Each of the three raters evaluated each of the 120 selected studies, presented in a randomised order.

One study was presented to the rater at a time. There was no time limitation imposed on the raters. They first reviewed the available information for the study, including:
\begin{itemize}
    \item All the CXRs from the study and its prior;
    \item The indication and history sections from the radiology reports of the study and its prior;
    \item The findings and impression sections from the radiologist report of the prior; and
    \item The elapsed time between the study and its prior.
\end{itemize}
Next, the rater would review the generated and radiologist reports. The source of each report was blinded and their order randomised. The rater then indicated their pairwise preference between the generated and radiologist reports, with the option to indicate no preference. If a preference was indicated, the rater would then select one or more reasons for their preference: 
\begin{itemize}
    \item Precision: ``It had fewer incorrect findings, impressions/diagnoses, and recommendations.''
    \item Recall: ``It captured more of the important findings,
impressions/diagnoses, and recommendations.''
    \item Readability: ``It was written in a clearer, more concise, or better
structured manner.''
\end{itemize}
The specification of the findings, impressions/diagnoses, and recommendations was based on the work of \cite{hartung_how_2020}, which describes radiology reports as structured clinical documents that primarily communicate imaging observations (findings), their clinical interpretation (impressions or diagnoses), and suggested clinical actions (recommendations). These three components collectively capture the principal informational content of a radiology report and therefore provide a practical basis for evaluating reports in terms of factual correctness (precision), completeness of clinically relevant information (recall), and clarity of communication (readability).

The primary outcome was categorical report preference (generated report, radiologist report, or no preference). Secondary data included the preference reasons (precision, recall, and readability), optional qualitative comments, and timestamps capturing study-level evaluation duration. Raters could pause and resume evaluations at their discretion, allowing completion across multiple sessions. All data were stored in a password-protected, encrypted database with access restricted to the primary investigator. All statistical analyses were conducted using R (version 4.5.2) \citep{r_core_team_r_2025}.

\subsection{Results \& discussion} \label{sec:results_retro}

\begin{figure*}[]
  \centering
  \includegraphics[width=\textwidth]{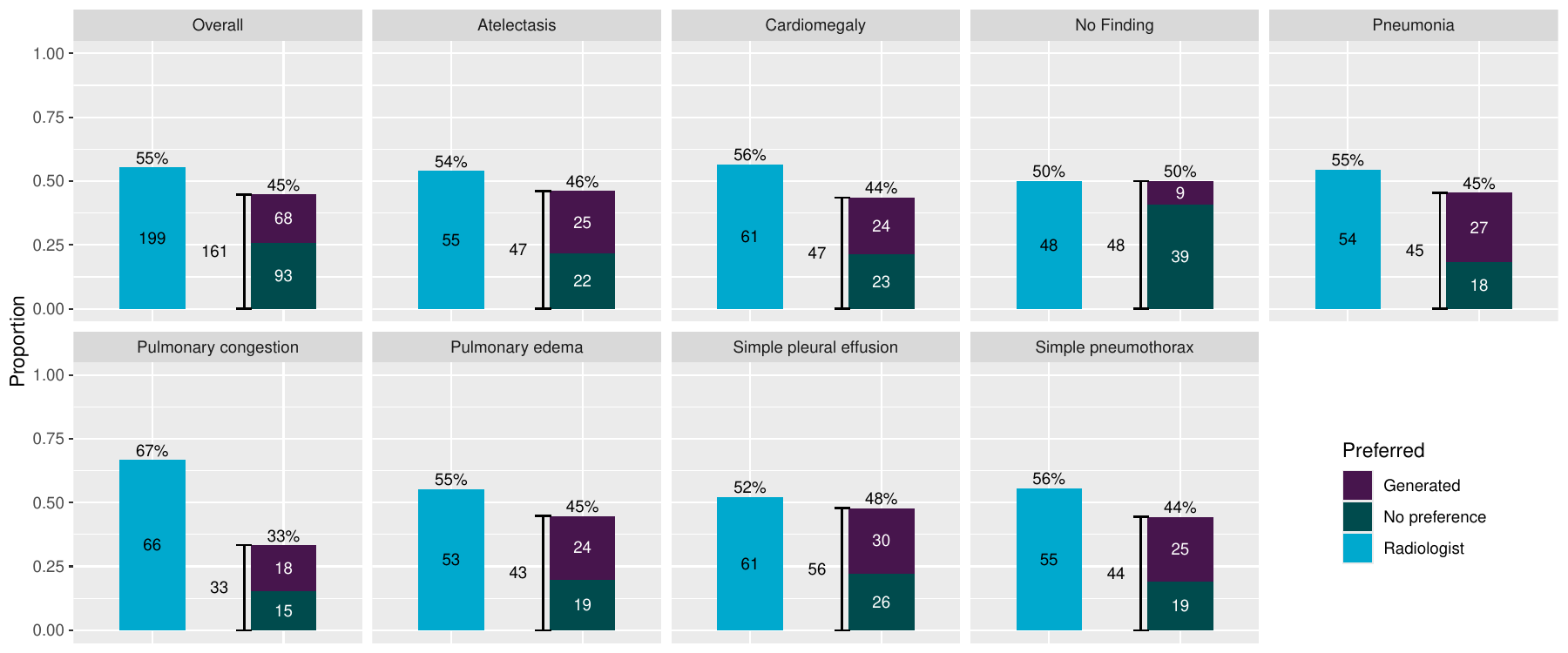}
  \caption{Preferences from three consultant radiologist raters for the 120 selected studies.}
  \label{fig:radiologist_preferences}
\end{figure*}

\begin{figure*}[]
  \centering
  \includegraphics[width=\textwidth]{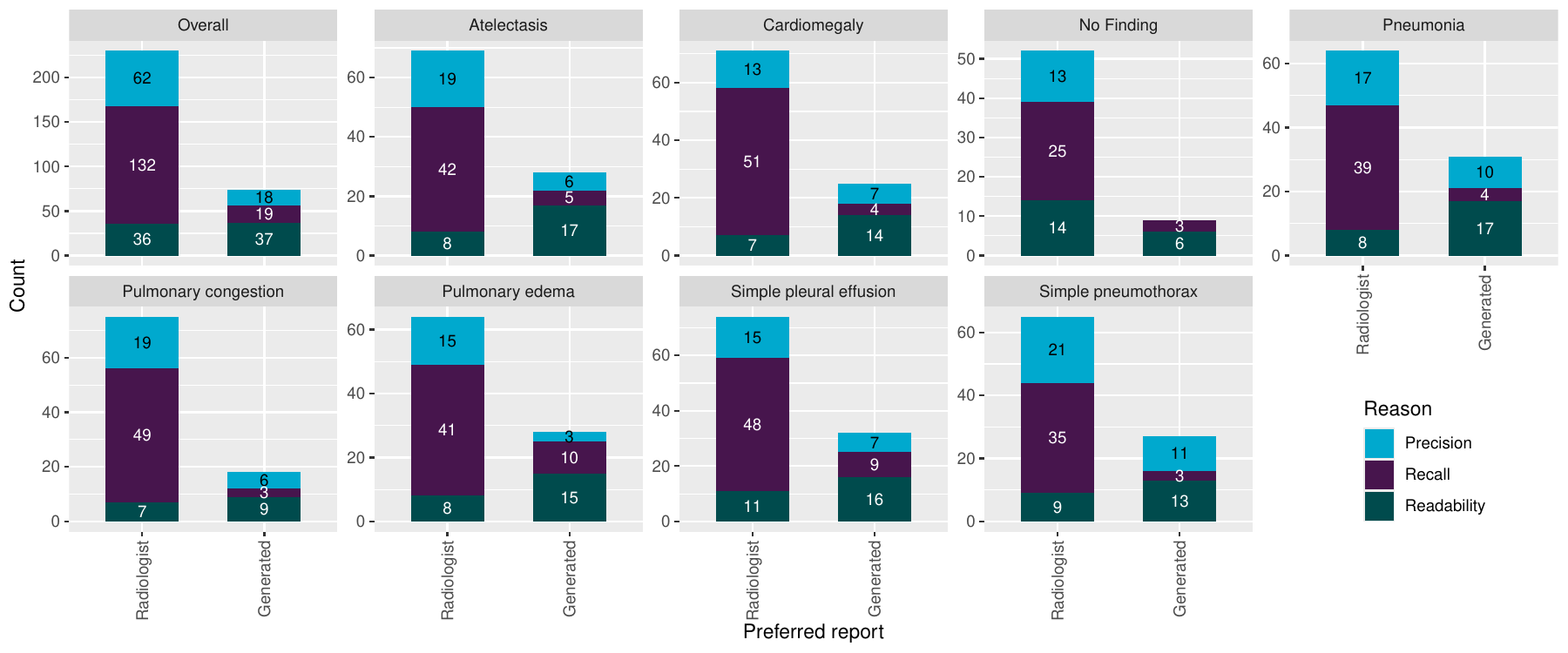}
  \caption{Reasons for the preferences from three consultant radiologist raters for the 120 selected studies.}
  \label{fig:radiologist_reasons}
\end{figure*}

\subsubsection{Preferences} \label{sec:results_pref}

The primary goal of the evaluation was to test whether generated reports were non-inferior to radiologist reports under a predefined acceptability endpoint. We formalise this using an acceptability endpoint defined as $P_{acc} = \mathbb{P}(Y = 1)$, where
\[
Y =
\begin{cases}
1 & \text{if generated report is preferred or no preference;} \\
0 & \text{if radiologist report is preferred.}
\end{cases}
\]
$P_{acc}$ is the proportion of studies in which the generated report is either preferred or judged equivalent to the radiologist report.

Following \cite{piaggio_reporting_2012}, non-inferiority is defined relative to a threshold of 50\%, corresponding to the setting in which generated reports are not judged inferior in at least half of the studies. Under this definition, the probability of a generated report being acceptable is equal to the probability of a radiologist report being preferred. We utilise the following one-sided hypothesis test:
\[
H_0: P_{acc} \le 0.5 - \delta \quad \text{vs.} \quad H_1: P_{acc} > 0.5 - \delta,
\]
where we set a conservative non-inferiority margin of $\delta=0$, corresponding to a strict criterion in which generated reports must achieve acceptability exceeding 50\%. 

Rejection of $H_0$ would indicate that generated reports are non-inferior to radiologist reports under this definition. While this criterion does not establish full clinical equivalence, if met, it would provide preliminary evidence supporting the feasibility of a CXR RRG model as an assistive tool for a radiologist-led workflow.

The radiologist preferences are shown in Figure \ref{fig:radiologist_preferences}. Overall, generated reports were deemed acceptable in 45\% of ratings; this did not significantly exceed the 50\% threshold required for non-inferiority ($p=0.051$, Table \ref{tab:pathology_counts}).

Acceptability was highest for studies with no findings ($P_{acc}=50\%$) and lowest for studies involving pulmonary congestion ($P_{acc}=33\%$). Binomial proportion tests (Table~\ref{tab:pathology_counts}) indicate that the difference between acceptable generated report and radiologist report preferences was statistically significant only for pulmonary congestion ($p=0.001$). 

Pulmonary congestion is challenging to detect as it manifests as a diffuse, low-contrast spectrum of vascular and interstitial changes. Its radiographic signs are subtle, non-specific, and often require relational assessment of vascular redistribution across lung zones, making interpretation dependent on both fine-grained detail and global context.

\begin{table}[]
\centering
\setlength{\tabcolsep}{4pt}
\small
\caption{Binomial proportion test of the rater's preferences. This assumes independence and therefore likely overestimates effective sample size. **Significant at $p \leq 0.01$.}
\label{tab:pathology_counts}

\rowcolors{2}{gray!15}{white}

\begin{tabular}{p{2cm}c>{\centering\arraybackslash}p{2cm}c}
\toprule
Finding & Radiologist & Generated or no preference & $p$ \\ \midrule
Overall & 199 & 161 & 0.051 \\
Atelectasis & 55  & 47 & 0.488 \\
Cardiomegaly & 61  & 47  & 0.211 \\
No Finding & 48  & 48  & 1.000 \\
Pneumonia & 54  & 45 & 0.422 \\
Pulmonary congestion  & 66  & 33  & 0.001** \\
Pulmonary edema & 53  & 43  & 0.358 \\
Simple pleural effusion & 61  & 56  & 0.712 \\
Simple pneumothorax & 55  & 44  & 0.315 \\ 
\bottomrule
\end{tabular}
\end{table}

As highlighted in Subsection \ref{sec:data}, a large proportion of the selected studies are complex. This means that the preferences in Figure \ref{fig:radiologist_preferences} are not isolated for each finding. To truly isolate the performance for each finding, studies would need to be selected with only that finding included.

\subsubsection{Reasons} \label{sec:results_reasons}

The reasons underlying rater preferences are summarised in Figure \ref{fig:radiologist_reasons}. Across all findings, radiologist reports were most frequently preferred due to higher recall, accounting for 57\% of all recorded reasons for radiologist report preference. Together with the overall majority preference for radiologist reports, this suggests that radiologist reports tended to capture a greater proportion of findings, impressions/diagnoses, and recommendations per study.

To a lesser extent, radiologist reports were preferred due to higher precision across all findings, accounting for 27\% of all recorded reasons for radiologist report preference. Combined with the overall preference for radiologist reports, this suggests that radiologist reports tended to contain fewer incorrect findings, impressions/diagnoses, and recommendations.

In contrast, when readability was the reason, generated reports tended to be preferred over radiologist reports, particularly in studies with abnormal findings. Overall, readability accounted for 50\% of all recorded reasons for generated report preference, suggesting that generated reports were more consistently written in a clearer, more concise, or better structured manner.

\begin{figure*}[]
  \centering
  \includegraphics[width=\textwidth]{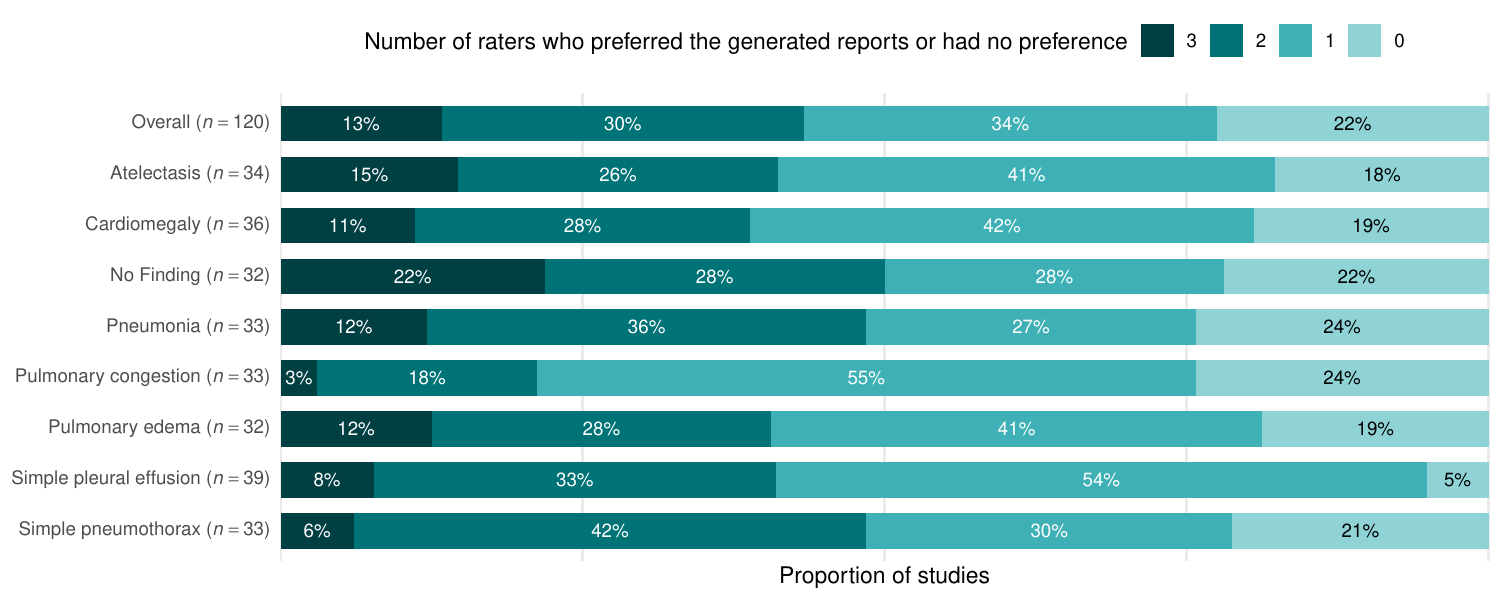}
  \caption{Proportion of inter-rater agreement of preferences between the three consultant radiologist raters for the 120 selected studies.}
  \label{fig:acceptance}
\end{figure*}


\subsubsection{Inter-rater agreement} \label{sec:inter}

The distribution of inter-rater agreement is shown in Figure \ref{fig:acceptance}. Overall, 43\% of generated reports were deemed acceptable---that is, they were preferred or equally preferred to radiologist reports---by a majority of raters (i.e., 2 or 3 raters). This distribution is similar to that observed in Figure 3 of \cite{tanno_collaboration_2025}, suggesting that comparable levels of inter-rater agreement have been reported in prior work. Reports with no findings had the highest majority acceptability (50\%), while pulmonary congestion had the lowest (21\%), consistent with previous observations. This indicates that majority acceptability is strongly modulated by finding, with generated reports performing comparatively better on normal studies, while remaining challenged by subtle, diffuse pathologies such as pulmonary congestion.


To account for agreement expected by chance, inter-rater agreement was quantified using Fleiss’ $\kappa$. The estimated overall agreement beyond chance across all raters was low ($\kappa = 0.16$), corresponding to slight agreement under conventional interpretation guidelines. Although $\kappa$ differed significantly from zero ($z = 4.14$, $p < 0.001$), the magnitude of $\kappa$ indicates that practical agreement between raters was weak. This level of agreement limits the interpretability of absolute preference rates and suggests substantial variability in rater preferences, reinforcing the need for larger sample sizes and rater pools to obtain stable estimates.

The pairwise agreement between each pair of raters is presented in Table \ref{tab:pairwise_agreement}. The highest agreement was observed between Raters B and C (53\%, $\kappa = 0.23$), corresponding to fair agreement and representing the only pair with statistically significant agreement beyond chance ($p = 0.001$). In contrast, agreement between Raters A and B (49\%, $\kappa = 0.10$) and between Raters A and C (48\%, $\kappa = 0.13$) was weaker and not statistically distinguishable from chance at the 0.05 significance level. Notably, Raters B and C had similar mean evaluation durations per study (212 and 197 seconds, respectively), whereas Rater A’s was substantially shorter (108 seconds), which may partially contribute to the lower agreement involving Rater A.\footnote{These estimates exclude durations above the $95^{th}$ percentile. The distribution of evaluation durations per study is provided in Appendix \ref{sec:study_eval_duration}.} This indicates that inter-rater agreement is not uniform, but varies depending on the raters being compared. This heterogeneity may reflect differences in factors such as evaluation duration, experience, or fatigue.

The low inter-rater agreement may also reflect that making preference judgements was inherently difficult when generated and radiologist reports were similar. The three examples in Figure~\ref{fig:visualisation_disagreement} suggest that low inter-rater agreement may partly reflect difficulty distinguishing highly similar generated and radiologist reports. Consistent with these examples, complete disagreement occurred in 8\% of studies, while only two radiologists agreed in 71\%, indicating that unanimous consensus was uncommon. Taken together, close semantic and linguistic similarity between generated and radiologist reports may render preference judgements ambiguous, contributing to low inter-rater agreement.

\begin{table}[]
\centering
\setlength{\tabcolsep}{4pt}
\small
\caption{Inter-rater agreement (Fleiss’ $\kappa$). **Significant at $p\leq0.01$. 
}
\label{tab:pairwise_agreement}
\rowcolors{2}{gray!15}{white}

\begin{tabular}{lcccc}
\toprule
Raters &Agreement& $\kappa$& $z$& $p$ \\ \midrule
Overall &29\%& 0.16 & 4.1 & $<0.001$** \\
A vs. B &49\%& 0.10& 1.5& 0.145 \\
A vs. C &48\%& 0.13& 1.9& 0.060 \\
B vs. C &53\%& 0.23& 3.3& 0.001** \\ \bottomrule\end{tabular}
\end{table}

\subsubsection{Statistical modelling and factor effects} \label{sec:effects}

To analyse how different factors---such as rater, reason, and finding---affect whether generated reports are judged \textit{acceptable} (i.e., preferred or considered equivalent to radiologist reports), we fitted a generalised linear model (GLM) with a binomial distribution. Odds ratios (ORs) represent the change in odds of acceptability relative to a baseline. The reference categories for the baseline were reason = readability, rater = A, and finding = \textit{No Finding} (OR = $0.98 \approx 1.0$ , $p=0.967$). Table~\ref{tab:glm_modified} summarises the estimated odds ratios and confidence intervals (CIs) for all predictors in the model. Table~\ref{tab:anova_deviance_full} identified significant interactions, with those with non-significant interactions omitted from Table~\ref{tab:glm_modified} (specifically, Reason $\times$ Finding and Rater $\times$ Finding). The key findings include:

\begin{itemize}
    \item \textbf{Reason dominates preference:} 
    Generated reports were substantially less likely to be preferred under precision (OR = 0.09, $p<0.001$) or recall (OR = 0.04, $p<0.001$) compared with readability. This suggests that current models do not yet consistently match radiologists in factual correctness or completeness.

    \item \textbf{Generated reports were more readable for abnormal studies:} Generated reports were more likely to be preferred for readability in abnormal findings, relative to normal studies, suggesting their primary advantage lies in linguistic quality.

    \item \textbf{Finding-specific effects are limited to readability:} No significant Reason $\times$ Finding interaction was observed in Table \ref{tab:anova_deviance_full}, indicating that differences across findings should not be interpreted for precision or recall and are primarily driven by readability (i.e., the normal versus abnormal difference in readability noted above). This may also reflect that the selected studies often contain multiple findings, limiting the ability to isolate finding-specific effects.

    \item \textbf{Rater variability is substantial:} Rater C was significantly less likely to judge generated reports acceptable than Rater A under readability (OR = 0.16, $p<0.001$). This further confirms that there is low inter-rater agreement.

    \item \textbf{Rater effects depend on the reason:} Significant interactions between Rater C and both precision (OR = 6.08, $p=0.005$) and recall (OR = 5.00, $p=0.010$) suggest that rater differences are not uniform, but vary depending on the reason for the preference.

    \item \textbf{No evidence of rater–finding dependence:} The absence of a significant Rater $\times$ Finding interaction in Table \ref{tab:anova_deviance_full} indicates that rater-specific tendencies were consistent across findings. This is somewhat counterintuitive given the low inter-rater agreement, suggesting that disagreement is driven primarily by other factors.

    \item \textbf{Effect size uncertainty:} Although several findings showed significant effects, wide confidence intervals indicate imprecise estimates due to limited samples per finding. A similar pattern is observed for the Reason $\times$ Rater interaction, suggesting that larger study sizes are needed for more stable estimates.

\end{itemize}




\begin{table}[]
\centering
\setlength{\tabcolsep}{6pt}
\small
\caption{GLM results for acceptable generated report preferences. The reference categories were reason = readability, rater = A, and finding = \textit{No Finding}. *Significant at $p\leq0.05$. **Significant at $p\leq0.01$.}
\label{tab:glm_modified}
\rowcolors{2}{gray!15}{white}

\begin{tabular}{lccc}
\toprule
Predictors & OR & 95\% CI & $p$ \\
\midrule
Readability & 0.98 & 0.39--2.33 & 0.967 \\
Precision & 0.09 & 0.03--0.23 & $<0.001$** \\
Recall & 0.04 & 0.01--0.10 & $<0.001$** \\ 
\hline
Rater B & 0.53 & 0.20--1.47 & 0.211 \\
Rater C & 0.16 & 0.07--0.36 & $<0.001$** \\
\hline
Atelectasis & 4.23 & 1.59--12.04 & 0.005* \\
Cardiomegaly & 4.15 & 1.56--11.83 & 0.006* \\
Pneumonia & 4.54 & 1.73--12.79 & 0.003* \\
Pulmonary congestion & 2.64 & 0.95--7.71 & 0.067 \\
Pulmonary edema & 4.56 & 1.71--13.01 & 0.003* \\
Simple pleural effusion & 4.53 & 1.76--12.59 & 0.002* \\
Simple pneumothorax & 4.44 & 1.70--12.51 & 0.003* \\
\hline
Precision $\times$ Rater B & 2.11 & 0.58--7.53 & 0.252 \\
Recall $\times$ Rater B & 1.36 & 0.38--4.80 & 0.631 \\
Precision $\times$ Rater C & 6.08 & 1.70--21.59 & 0.005* \\
Recall $\times$ Rater C & 5.00 & 1.45--16.93 & 0.010* \\
\bottomrule
\end{tabular}
\end{table}

\begin{table}[]
\centering
\caption{Analysis of deviance (Type I likelihood ratio tests) for the fitted GLM. Columns: df = degrees of freedom associated with each term; $\Delta D$ = change in deviance (likelihood ratio statistic) when the term is added to the model; df$_{\mathrm{res}}$ = residual degrees of freedom after including the term; $D_{\mathrm{res}}$ = residual deviance of the model after including the term; $p$ = p-value from a likelihood ratio test, where $\Delta D \sim \chi^2_{\mathrm{df}}$.}
\label{tab:anova_deviance_full}
\rowcolors{2}{gray!15}{white}

\begin{tabular}{lccccc}
\toprule
Term & df & $\Delta D$ & df$_{\mathrm{res}}$ & $D_{\mathrm{res}}$ & $p$ \\
\midrule
Reason                    & 2  & 141 & 729 & 713 & $<0.001$ \\
Rater                     & 2  & 10   & 727 & 704 & 0.008 \\
Findings               & 7  & 16  & 720 & 688 & 0.025 \\
Reason $\times$ Rater              & 4  & 17  & 716 & 670 & 0.002 \\
Reason $\times$ Finding        & 14 & 17  & 702 & 653 & 0.270 \\
Rater $\times$ Finding         & 14 & 8 & 688 & 645 & 0.883 \\
\bottomrule
\end{tabular}
\end{table}




\subsubsection{Power analysis}

\paragraph{\textbf{Power of the current evaluation:}}
\textit{A priori} power analysis was not conducted because reliable estimates of key parameters---particularly effect size (e.g., $P_{acc}$ relative to 0.5) and inter-rater variability---are not available in the existing CXR RRG literature. Consequently, any \textit{a priori} analysis would have required strong and potentially unjustified assumptions. Instead, this work was designed as a pilot, with post hoc power analysis used to quantify its limitations and guide the design of future adequately powered evaluations.

\begin{figure}[]
  \centering
  \includegraphics[width=\columnwidth]{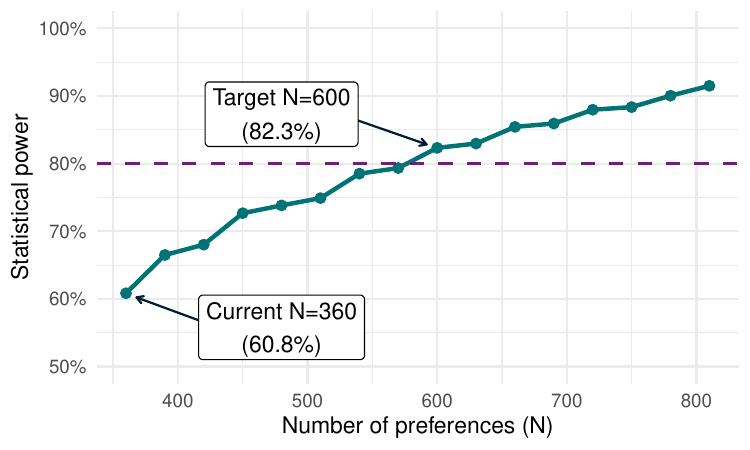}
  \caption{Binomial power analysis of the raters preferences.}
  \label{fig:BinPowerCurve}
\end{figure}

The post hoc power analysis was conducted over the 360 preferences from the three raters over the 120 selected studies (``Overall'' in Figure \ref{fig:radiologist_preferences} and Table \ref{tab:pathology_counts}). As shown in Figure \ref{fig:BinPowerCurve}, a one-sided binomial proportion test against a null hypothesis ($P = 0.5$) indicates that the current sample of 360 preferences yields a power of 60.8\%.

\paragraph{\textbf{Powering future evaluations:}}

In Appendix \ref{sec:scenarios}, we present a power analysis to inform the design of future evaluations. Under a near-equivalence setting (preference $\approx 50\%$ for precision and recall), the current pilot configuration is underpowered (approximately 57\%). Adequate power ($\geq 80\%$) can be achieved either by increasing the number of raters (e.g., to six, achieving 88.4\% power for precision and 93.6\% for recall) or by increasing the total number of ratings (approximately 600 under a binomial formulation yields 82.3\% power; 720 yields 87.9\%; Figure \ref{fig:BinPowerCurve}). These results provide practical guidance for selecting sample sizes in future evaluations, depending on whether expansion is more feasible along the rater or study-count dimension.



\subsubsection{Qualitative examples of preferences}

\begin{figure*}[]
  \centering
  \includegraphics[width=\textwidth]{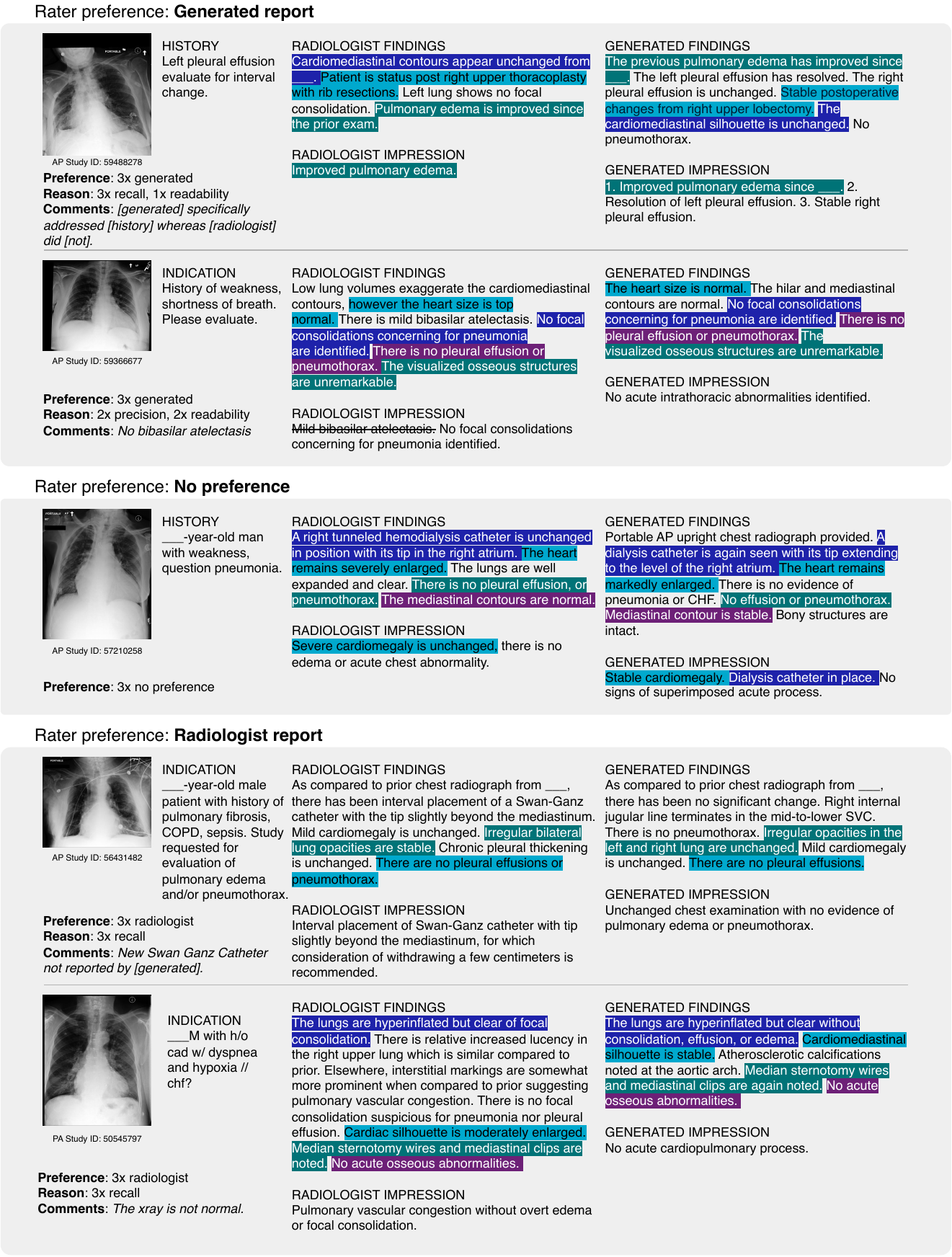}
  \caption{Generated and radiologist reports for studies with different preferences. Only a frontal view of each study is shown. Strikethrough indicates an incorrect finding relative to the preferred report. Corresponding sentences are highlighted with matching colours across the radiologist and generated reports to aid comparison. Square brackets indicate modifications to the original text.}
  \label{fig:visualisation}
\end{figure*}

Figure \ref{fig:visualisation} presents paired generated and radiologist reports for studies where the overall preference was for the generated report, the radiologist report, or no preference.

The first and second rows depict studies where the generated report was preferred. In the first row, the generated report explicitly addresses interval changes described in the history (resolution of the left pleural effusion), whereas the radiologist report omits this detail. In the second example, the generated report avoids an incorrect finding (bibasilar atelectasis) present in the radiologist report, leading to higher perceived precision.

The third row is an example where no preference is expressed. In this case, both reports convey largely equivalent findings and impressions. Differences are primarily stylistic, with both adequately capturing the key clinical information, leading to indifference among raters.

Finally, the fourth and fifth rows are studies where the radiologist report was preferred. These examples highlight common failure modes of the generated reports, particularly in recall. In the fourth row, the generated report fails to identify the interval placement of a Swan–Ganz catheter and the recommendation for repositioning. In the fifth example, the generated report incorrectly concludes that there is no acute cardiopulmonary process, missing the pulmonary vascular congestion.

These examples demonstrate that generated reports can match or exceed radiologist reports in specific cases. However, the broader results indicate that this performance is not yet consistent, and improving reliability---particularly in recall---is necessary for practical use.

\section{Limitations}

Several limitations should be considered when interpreting the findings of this work.

First, the model was trained exclusively on datasets derived from US healthcare institutions (MIMIC-CXR, CheXpert Plus, and ReXgradient). Radiology reporting style, clinical practice patterns, patient demographics, and imaging protocols can vary substantially across institutions and countries. As a result, the performance of CXRMate-2 may not generalise directly to other clinical settings without additional adaptation or evaluation on locally representative datasets.

Second, the qualitative retrospective evaluation was designed as a pilot and therefore has limited statistical power. Although 360 individual preferences were collected across three consultant radiologists and 120 studies, the resulting power for detecting the observed effect size was approximately 60.8\%. Consequently, the absence of statistically significant differences in some comparisons should be interpreted cautiously, as the study may be underpowered to detect smaller differences between generated and radiologist reports. The pilot design nevertheless provides useful estimates of effect sizes, inter-rater variability, and sample size requirements that can inform the design of larger and more adequately powered evaluations.

Third, inter-rater agreement among the radiologists was relatively low, indicating only slight agreement beyond chance. While variability in interpretation is expected in radiology, low agreement introduces uncertainty in preference-based evaluations and reduces statistical sensitivity. This may also have reflected studies where generated and radiologist reports were highly similar, making preference judgements inherently ambiguous. Future studies may benefit from larger rater pools.

Fourth, the qualitative analysis indicates that generated reports frequently underperform radiologist reports in terms of recall and, to a lesser extent, precision. This suggests that---although the model often produces fluent and readable reports---it may omit clinically important observations or include occasional inaccuracies. Improving recall---while also addressing precision---remains an important direction for future work.

Fifth, model performance appears to vary across finding. In particular, the evaluation revealed substantially lower acceptability for studies involving pulmonary congestion. This likely reflects the inherent difficulty of detecting subtle, diffuse radiographic patterns that require both fine-grained visual detail and contextual interpretation across lung regions. More broadly, the analysis suggests that certain findings may remain challenging for current CXR RRG models, highlighting the need for targeted improvements in visual representation learning. However, these finding-specific analyses were conducted on studies that often contain multiple co-occurring findings, and therefore do not isolate single-finding performance. As a result, the observed differences should be interpreted cautiously, as they may be influenced by interactions between findings within the same study. Nevertheless, evaluation on complex, multi-finding studies should be included as a key metric for real-world clinical utility.

\section{Conclusion} 

We presented CXRMate-2, a unified framework for chest X-ray (CXR) radiology report generation that enables tractable reinforcement learning (RL) on modest compute resources via structured multimodal temporal embeddings for scalable visual, textual, and temporal conditioning and a query-based Transformer adapter (Q-Adapter)---for efficient high-resolution visual feature compression. We show that while these components substantially reduce self-attention computational complexity, they also improve CXR RRG performance. These components enable tractable group relative policy optimisation (GRPO), which yields improved performance over previous RL algorithms. Furthermore, our proposed composite reward function, which incorporates RaTEScore, further improves performance. Overall, CXRMate-2 achieves state-of-the-art performance across a broad set of automated metrics on MIMIC-CXR, CheXpert Plus, and ReXgradient, with statistically significant gains over strong baselines, including improvements of 11.2\% and 24.4\% in GREEN and RadGraph-XL, respectively, on MIMIC-CXR relative to MedGemma 1.5 (4B).

In a blinded, qualitative retrospective evaluation with consultant radiologists, generated reports were deemed acceptable---defined as preferred or equivalent to radiologist reports---in 45\% of ratings, with no statistically significant difference observed for seven of the eight targeted findings. A significant deficit was observed only for pulmonary congestion, highlighting persistent challenges in modelling subtle findings. Radiologist preference was driven primarily by higher recall, whereas generated reports were more often favoured for readability, particularly in abnormal studies. Inter-rater agreement was low, indicating substantial variability in rater judgement, potentially reflecting the inherent difficulty of distinguishing between similar reports, and emphasising the need for larger, better-powered evaluations. Together, these results indicate that CXRMate-2 approaches radiologist-level performance, with remaining limitations primarily in recall and subtle findings.

With further development, CXR RRG models may be well positioned to be prospectively evaluated as assistive tools within radiologist-led workflows rather than as stand-alone systems. Potential use cases include first-reader draft generation, second-reader quality assurance, triage based on study severity, and predictive text to accelerate report writing.

\bibliographystyle{cas-model2-names}

\bibliography{25_cxrmate2}



\appendix
\section*{Appendix}

\setcounter{figure}{0}
\renewcommand{\thefigure}{A\arabic{figure}}

\setcounter{table}{0}
\renewcommand{\thetable}{A\arabic{table}}

\setcounter{lstlisting}{0}
\renewcommand{\thelstlisting}{A\arabic{lstlisting}}

\begin{figure*}[]
\centering
\begin{tcolorbox}[
  colback=gray!5,
  colframe=gray!60,
  boxrule=0.5pt,
  arc=2pt,
  left=6pt,
  right=6pt,
  top=4pt,
  bottom=4pt,
  width=\textwidth
]
\begin{Verbatim}[breaklines=true, breakanywhere=true, fontsize=\tiny]
  ROLE
  You are an autorater determining the validity of a given chest X-ray (CXR) study.
  The criteria for validity is as follows:
  - The prior study must be the *immediate prior* study to the current CXR study. 
  - If no prior study is given, the study must be the first study for the patient.
  - The current study must not reference any imaging studies from other modalities.

  INPUTS
  - <CURRENT_STUDY_RADIOLOGY_REPORT>: the radiology report of the current study.
  - <PRIOR_STUDY_RADIOLOGY_REPORT>: the radiology report of the prior study.
  - <CURRENT_STUDY_DATETIME>: the date and time of acquisition of the current study in ISO 8601 format.
  - <PRIOR_STUDY_DATETIME>: the date and time of acquisition of the prior study in ISO 8601 format.
  - <TIME_DELTA>: the elapsed time between acquisition of the current and prior studies in ISO 8601 duration format.
  - If there is no prior study, <PRIOR_STUDY_RADIOLOGY_REPORT>, <PRIOR_STUDY_DATETIME>, and <TIME_DELTA> will be marked as N/A.
 
  BACKGROUND
  - The studies are from the MIMIC-CXR dataset.
  - Each study is a chest X-ray with an associated radiology report.
  - The radiology reports have been de-identified.
  - Because of gaps in patient history, the provided prior study may not be the true immediate prior study.

  GOAL
  - Return **1** if there is *no evidence* contradicting that the given prior study is the immediate prior study.
  - Return **0** if there is *any evidence* indicating that the given prior study is **not** the immediate prior study.
  - Return **0** if the current study references any imaging studies from other modalities. 

  GOAL IF NO PRIOR
  - Return **1** if there is *no evidence contradicting* that the current study is the first study for the patient.
  - Return **0** if there is *any evidence* indicating that the current study is **not** the first study for the patient.
  - Return **0** if the current study references clinical information not contained in either the current or prior radiology report. 

  EXAMPLE EVIDENCE FOR label: 0
  - <CURRENT_STUDY_RADIOLOGY_REPORT> references any imaging studies from other modalities, such as CT, MRI, ultrasound, or PET scans.

  If there is a prior study:
  - <CURRENT_STUDY_RADIOLOGY_REPORT> references a time interval (e.g., "No significant interval change in the past 24 hours") that conflicts with <TIME_DELTA> (e.g., 4 days).
  - <CURRENT_STUDY_RADIOLOGY_REPORT> references a prior study at a specific time (e.g., "compared to study at 2:29 p.m.") that does not match <PRIOR_STUDY_DATETIME>.
  - <CURRENT_STUDY_RADIOLOGY_REPORT> references an ambiguous time (e.g., "since earlier in the day") inconsistent with <PRIOR_STUDY_DATETIME>.
  - <CURRENT_STUDY_RADIOLOGY_REPORT> mentions unchanged findings (e.g., "Moderate left basal atelectasis is unchanged") that contradict findings in <PRIOR_STUDY_RADIOLOGY_REPORT> (e.g., "No atelectasis").
  - <CURRENT_STUDY_RADIOLOGY_REPORT> notes improvement or worsening of findings (e.g., "Slight improvement of left lower lobe opacity") that were not mentioned in <PRIOR_STUDY_RADIOLOGY_REPORT>.
  - <CURRENT_STUDY_RADIOLOGY_REPORT> references prior findings (e.g., "cardiac silhouette increased from prior exam") that are not present in <PRIOR_STUDY_RADIOLOGY_REPORT>.

  If there is no prior study:
    - <CURRENT_STUDY_RADIOLOGY_REPORT> makes any reference to a previous imaging study or time comparison, such as:
      - Mentioning a prior or earlier study (e.g., "compared to prior," "since previous exam," "no interval change").
      - Stating a time interval implying a prior study (e.g., "over the past 24 hours," "unchanged from earlier today").
      - Referring to findings as improved, worsened, or unchanged relative to an unstated earlier study.

  NOTES
  - The study is still valid if <CURRENT_STUDY_RADIOLOGY_REPORT> describes new or more specific findings.
  - Do not label: 0 merely because:
    - <CURRENT_STUDY_RADIOLOGY_REPORT> adds or omits minor details present in the prior.
    -	<CURRENT_STUDY_RADIOLOGY_REPORT> rephrases or paraphrases findings from <PRIOR_STUDY_RADIOLOGY_REPORT> without changing their meaning.

  OUTPUT (STRICT)
  Return ONE line of YAML only, with no explanation or extra text:
  label: 1 or 0
  reason: |-
    A short sentence (less than or equal to 10 words) containing your reasoning for the label.
\end{Verbatim}
\end{tcolorbox}
\caption{Prompt for Gemini 2.5 Flash to determine if a study and its reference to priors was valid.}
\label{fig:prior_study_validation_prompt}
\end{figure*}

\begin{figure*}[]
  \centering
  \includegraphics[width=\textwidth]{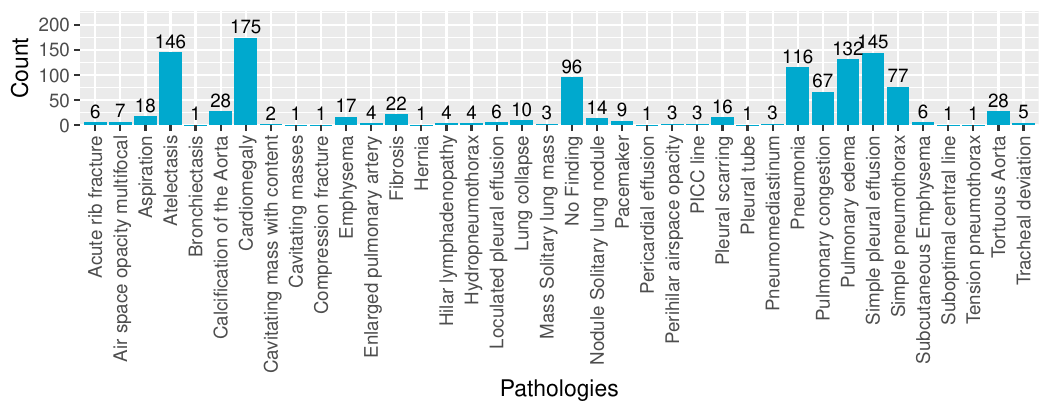}
  \caption{Findings for the 504 studies not excluded by the autorater described in Section \ref{sec:selection}.}
  \label{fig:pathology_distribution_valid}
\end{figure*}

\begin{figure}[]
  \centering
  \includegraphics[width=\columnwidth]{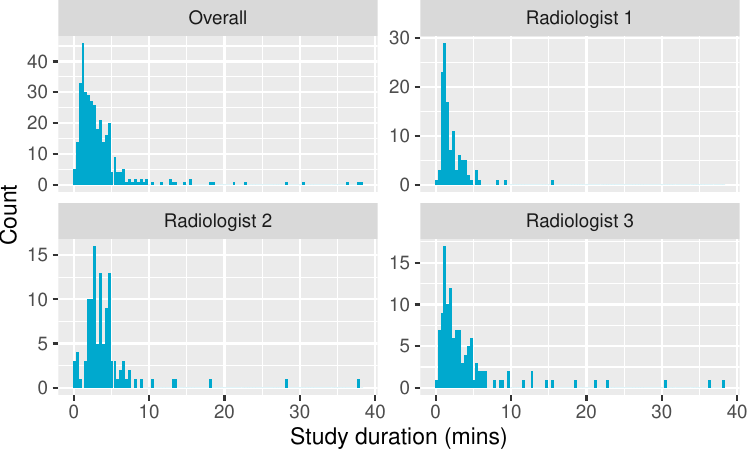}
  \caption{Distribution of evaluation durations per study, measured from initial presentation to final submission, with all viewing intervals summed to account for re-evaluations.}
  \label{fig:duration}
\end{figure}

\begin{figure*}[]
  \centering
  \includegraphics[width=\textwidth]{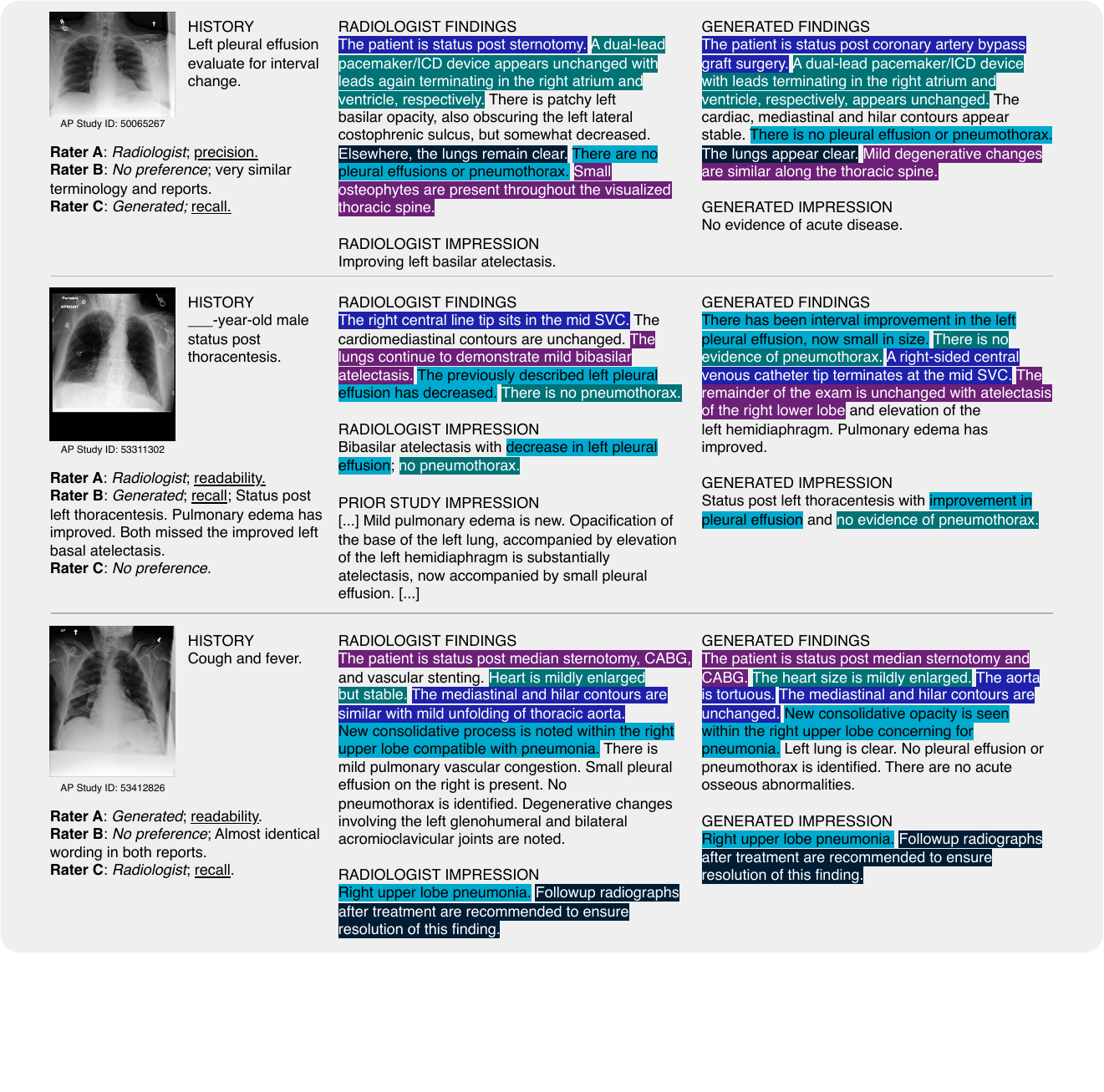}
  \caption{Generated and radiologist reports for studies with complete disagreement among raters. Only a frontal view of each study is shown. Corresponding sentences are highlighted with matching colours across the radiologist and generated reports to aid comparison. Square brackets indicate modifications to the original text.}
  \label{fig:visualisation_disagreement}
\end{figure*}

\section{Autorater prompt}

The prompt for the autorater described in Subsection \ref{sec:selection} is presented in Figure \ref{fig:prior_study_validation_prompt}.

\section{Valid studies}

The distribution of the findings in the studies deemed valid by the autorater are presented in Figure \ref{fig:pathology_distribution_valid}.

\section{Evaluation platform UI design} \label{sec:mockup}

Figure \ref{fig:mockup} illustrates a mock-up of the evaluation platform’s user interface (UI). The UI presented the rater with two panels such that the study was shown on the right and, if present, the  prior was shown on the left for comparison. Two viewers were available for displaying CXRs, one for the study and another for the prior. The viewers rendered DICOM images with interactive functionality, including zooming, panning, rotating, levelling, and inverting. Window level and width could be adjusted via mouse interaction or reset using a menu option. The viewers were available side-by-side in a panel, or each viewer was available as a pop-up. The raters were instructed to place the pop-ups for the two viewers side-by-side on a diagnostic medical display.

The UI presented thumbnails of all the CXRs for both the study and its prior (along with their respective views), allowing selection of specific CXRs for the two viewers. The viewers were synchronised with the main application to ensure the correct CXRs were shown when moving to the next or previous study, or when a different CXR was selected via the thumbnails. The viewers were implemented using Kitware VTK.js, a Javascript version of the C++ VTK library \citep{schroeder_visualization_2006}. The default photometric interpretation of most of the DICOMs was MONOCHROME2 so CXRs with MONOCHROME1 were inverted for correct visualisation. All DICOMs from the 120 selected studies, including their priors, used a linear values-of-interest look-up table (VOI LUT); therefore, windowing was applied to all DICOMs using a linear function defined by the window centre and width from the metadata.

If available, the referral for the study and its prior were presented, including the indication and history sections from the radiologist reports. If available, the findings and impression sections of the prior radiologist report were also presented. A generated and radiologist report for the current study was presented, with their order randomised. The preferred report, along with the reason or reasons for the preference could then be selected. If `No preference' was selected, then the ability to select a reason was disabled. The rater could also optionally leave a comment. The next study could be evaluated by clicking `Submit', or the previous study could be re-evaluated by clicking `Go back', where the previous inputs were maintained.

\section{Study evaluation duration} \label{sec:study_eval_duration}

The distribution of the evaluation duration per study is presented in Figure \ref{fig:duration}. The mean duration was 2.8 minutes when excluding durations above the $95^{th}$ percentile. This can help to estimate evaluation durations in future investigations.

\section{Disagreement visualisation}

Figure \ref{fig:visualisation_disagreement} presents three studies where the raters completely disagreed with one another.

\section{Power analysis for different scenarios} \label{sec:scenarios}

A power analysis was conducted to determine whether the current design of the qualitative retrospective evaluation is capable of detecting realistic improvements in model performance, and to inform the design of future evaluations. Generated report acceptability was modelled using a generalised linear mixed model (GLMM) with a binomial distribution, with \textit{reason} as a fixed effect (readability as the reference) and random effects for raters and findings to account for clustering. Power simulations assumed an intra-class correlation coefficient (ICC) of 0.033, reflecting clustering by rater and finding. To characterise sensitivity across different performance gaps, three effect-size scenarios were considered (\textit{easy}, \textit{moderate}, and \textit{hard}), representing progressively smaller differences between generated and radiologist reports.

These scenarios are defined quantitatively in Table~\ref{tab:power_effect_size}. The \textit{easy} scenario ($\beta=-1.10$, $\mathrm{OR}=0.33$) corresponds to a large performance gap, where the probability of a generated report being acceptable is 35.5\%. The \textit{moderate} scenario ($\beta=-0.70$, $\mathrm{OR}=0.50$) represents a smaller but still meaningful gap, with an acceptability probability of 45.1\%. The \textit{hard} scenario ($\beta=-0.50$, $\mathrm{OR}=0.61$) reflects near-equivalence between generated and radiologist reports, with an acceptability probability of approximately 50.1\%, making differences difficult to detect statistically. 

The current evaluation included 24 clusters (three raters $\times$ eight findings). Under this configuration, statistical power is limited: while large performance differences can be detected reliably, sensitivity decreases markedly as the gap between generated and radiologist reports narrows. In the most challenging setting---where generated reports approach radiologist-level performance---the estimated power is approximately 57\%, indicating that the current design is underpowered to detect subtle differences.

To address this, we evaluated an expanded design with six raters, yielding 48 clusters while retaining the same eight findings. The resulting sensitivity analysis is summarised in Table~\ref{tab:power_effect_size}. Under this expanded design, power is high across all scenarios. Even in the \textit{hard} scenario ($\beta=-0.50$, $\mathrm{OR}=0.61$)---which reflects near-equivalence between generated and radiologist reports (success probability $\approx 50\%$)---power remains high at 88.4\% for precision and 93.6\% for recall. These results indicate that the expanded design maintains strong sensitivity even when differences between generated and radiologist reports are small.

\begin{figure}[]
  \centering
  \includegraphics[width=\columnwidth]{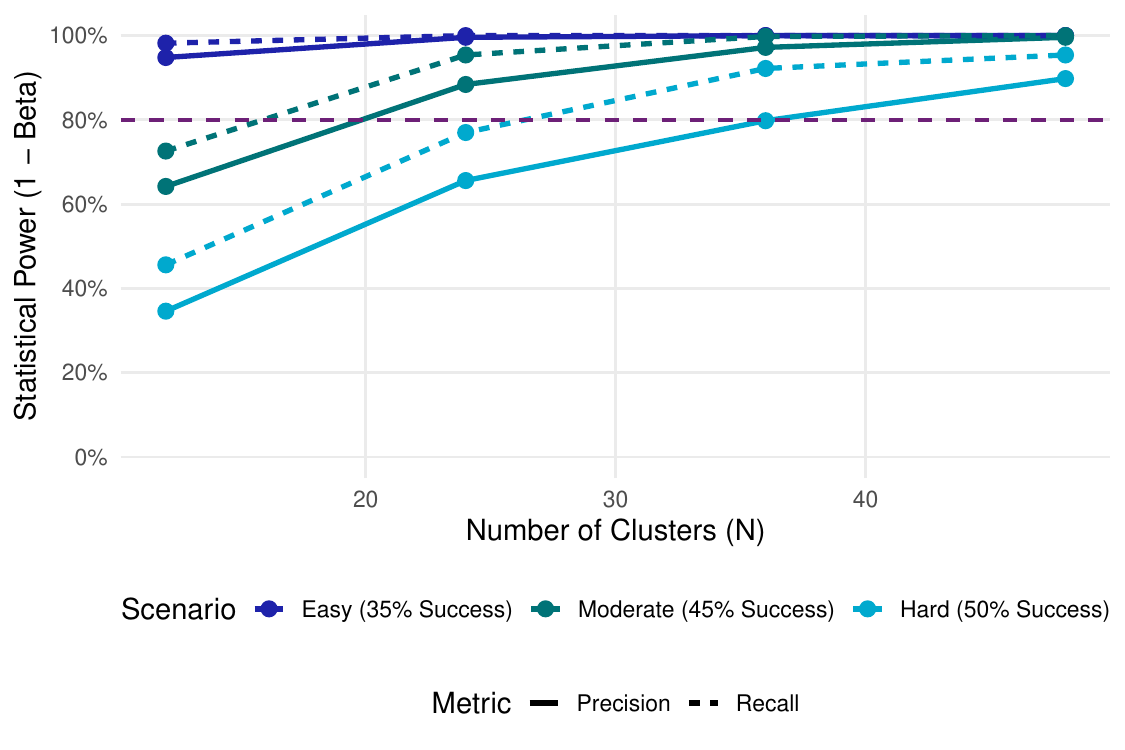}
  \caption{Power curves by metric (precision and recall) and predictive scenario. Purple dotted line indicates 80\% power threshold.}
  \label{fig:PowerCurve}
\end{figure}

\begin{figure*}[]
  \centering
  \includegraphics[width=\textwidth]{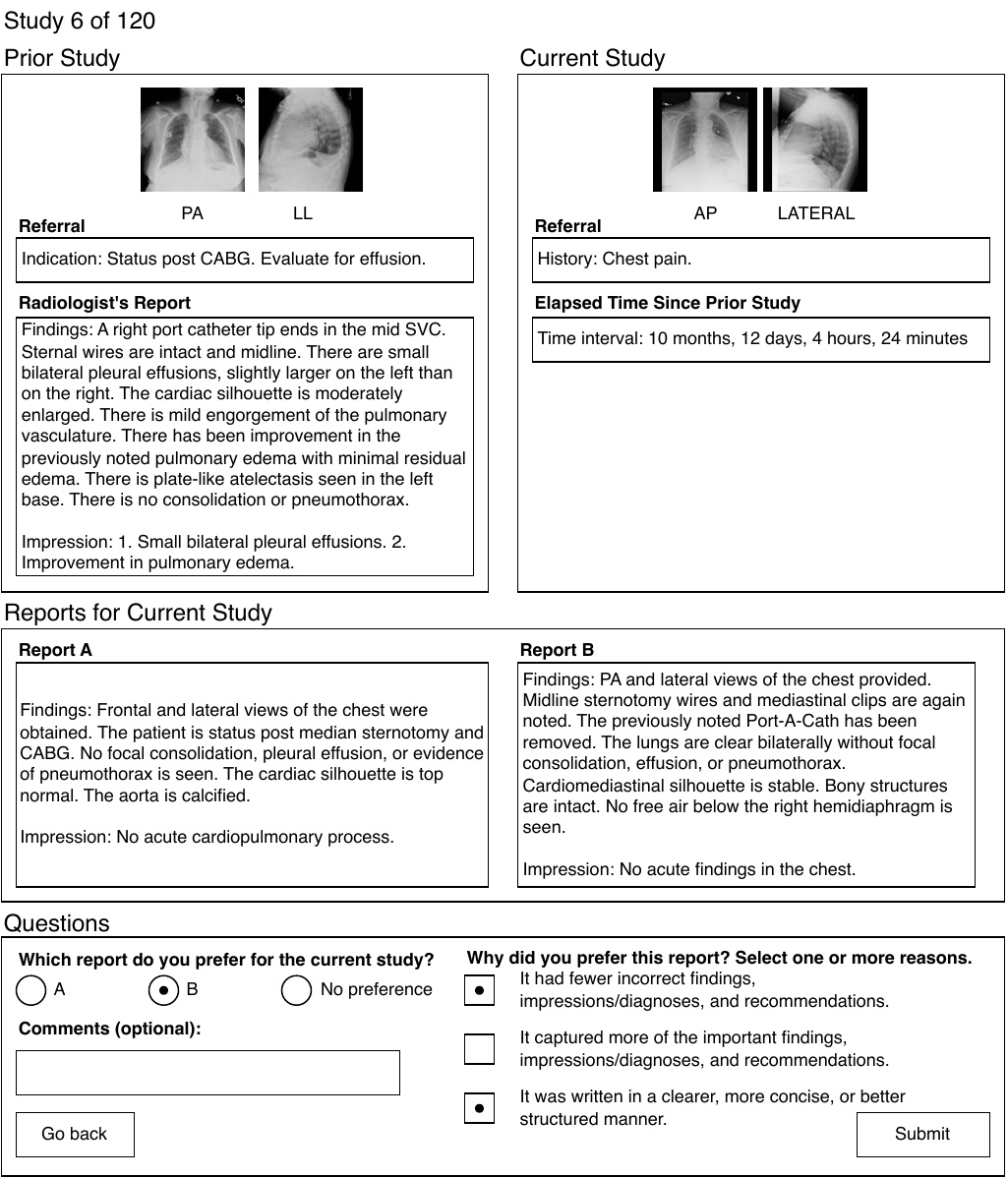}
    \caption{Mock-up of the user interface of the evaluation platform. Two viewers were available in separate windows: one displaying the selected CXR from the current study and the other displaying the selected CXR from the prior study, enabling side-by-side comparison.}
  \label{fig:mockup}
\end{figure*}

This behaviour is further illustrated in Figure~\ref{fig:PowerCurve}, which shows how statistical power increases as the number of clusters grows under each effect-size scenario. Each point in the curves is estimated from 500 simulations, and the figure highlights that the transition from 24 to 48 clusters moves the study from a low-power regime---particularly for the \textit{hard} scenario---to one that consistently exceeds the conventional 80\% power threshold.

These findings are consistent with the binomial power analysis presented in Subsection~\ref{sec:results_pref}. Under that formulation, increasing the total number of ratings to approximately 720---by evaluating more studies---yields a comparable power of 87.9\%. Overall, the results indicate that the current pilot design is sufficient for detecting large effects but lacks sensitivity for smaller differences. The agreement between the GLMM and binomial power analyses further suggests that adequate power can be achieved either by increasing the number of raters or by increasing the number of evaluated studies in future investigations.

\begin{table}[]
\centering
\setlength{\tabcolsep}{4pt}
\small
\caption{Sensitivity analysis: effect size and power for $N=48$ clusters.}
\label{tab:power_effect_size}
\rowcolors{2}{gray!15}{white}

\begin{tabular}{llcccc}
\hline
Scenario & Metric & $\beta$ & OR & Prob. (\%) & Power \\ \hline
Easy     & Precision & -1.10 & 0.33 & 35.5\% & 100.0\% \\
         & Recall    & -1.10 & 0.33 & 35.5\% & 100.0\% \\ \hline
Moderate & Precision & -0.70 & 0.50 & 45.1\% & 99.2\%  \\
         & Recall    & -0.70 & 0.50 & 45.1\% & 99.8\%  \\ \hline
Hard     & Precision & -0.50 & 0.61 & 50.1\% & 88.4\%  \\
         & Recall    & -0.50 & 0.61 & 50.1\% & 93.6\%  \\ \hline
\end{tabular}
\end{table}

\section{Normalising CXRs for CXRMate-2}

The following normalisation was applied only to the CXRs provided to CXRMate-2 and not to the CXRs presented to the raters. The CXRs in PNG and JPG formats from MIMIC-CXR-JPG, CheXpert Plus, and ReXgradient, which have an 8-bit pixel depth, were normalised using Listing \ref{lst:mimic_cxr_norm}.

\begin{lstlisting}[style=python, caption={MIMIC-CXR image normalisation and histogram equalisation.}, label={lst:mimic_cxr_norm}]label={lst:mimic_cxr_norm}
import cv2
import numpy as np
from PIL import Image

def mimic_cxr_normalisation(image: Image.Image) -> Image.Image:

    image_np = np.array(
        image.convert('L'), 
        dtype=np.float32
    )
    
    if image_np.ndim != 2:
        raise ValueError(f'Expected 2D grayscale image, got shape {image_np.shape}.')

    min_val = float(image_np.min())
    max_val = float(image_np.max())
    denom = max_val - min_val
    if denom == 0:
        raise ValueError(
            f'Cannot normalise image with zero dynamic range (min=max={min_val}).'
        )

    image_np = (image_np - min_val) / denom
    image_uint8 = (image_np * 255).astype(np.uint8)
    image_eq = cv2.equalizeHist(image_uint8)
    return Image.fromarray(image_eq)
\end{lstlisting}

\section{Benchmark models} \label{sec:benchmark_models} 

CXRMate-2 was benchmarked against the following models:

\begin{itemize}
    \item \textbf{EMNLI:} Reports were generated following \url{https://github.com/ysmiura/ifcc} \citep{miura_improving_2021}.
    \item \textbf{CXRMate:} Reports were generated following \url{https://huggingface.co/aehrc/cxrmate} \citep{nicolson_longitudinal_2024}.
    \item \textbf{CXRMate-RRG24:} Reports were generated following \url{https://huggingface.co/aehrc/cxrmate-rrg24} \citep{nicolson_e-health_2024}.
    \item \textbf{MAIRA-2:} Reports were generated following \url{https://huggingface.co/microsoft/maira-2} \citep{bannur_maira-2_2024}.
    \item \textbf{Libra:} Reports were generated following \url{https://github.com/X-iZhang/Libra} \citep{zhang_libra_2025}.
    \item \textbf{CXRMate-ED:} Reports were generated following \url{https://huggingface.co/aehrc/cxrmate-ed} \citep{nicolson_impact_2025}.
    \item \textbf{MedGemma (4B):} Reports were generated following \url{https://huggingface.co/google/medgemma-4b-it}, where all CXRs for a study, as well as the indication, history, and technique sections were included in the prompt, and the model was instructed to generate the findings section \citep{sellergren_medgemma_2025}.
    \item \textbf{MLRG:} The generated reports were obtained from \url{https://github.com/mk-runner/MLRG} \citep{liu_enhanced_2025}.
    \item \textbf{MedVersa:} Reports were generated following \url{https://huggingface.co/hyzhou/MedVersa} \citep{zhou_medversa_2026}.
    \item \textbf{PriorRG:} The generated reports were obtained from \url{https://github.com/mk-runner/PriorRG} \citep{liu_priorrg_2026}.
    \item \textbf{DeepMedix-R1:} Reports were generated following \url{https://huggingface.co/Qika/DeepMedix-R1} \citep{lin_toward_2026}.
    \item \textbf{MedGemma 1.5 (4B):} Reports were generated following \url{https://huggingface.co/google/medgemma-1.5-4b-it}, where all CXRs for a study, as well as the indication, history, and technique sections---and the CXRs, findings section, and impression section from the prior---were included in the prompt, and the model was instructed to generate the findings and impression section \citep{sellergren_medgemma_2026}.

\end{itemize}

\end{document}